\documentclass{article}

\PassOptionsToPackage{numbers,compress,sort}{natbib}
\usepackage{natbib}
    \usepackage[final]{neurips_2025}

\usepackage[utf8]{inputenc} % allow utf-8 input
\usepackage[T1]{fontenc}    % use 8-bit T1 fonts
\usepackage[x11names,table,xcdraw]{xcolor} 
\definecolor{citecolor}{rgb}{0.18, 0.57, 0.83}
\definecolor{linkcolor}{rgb}{0.90, 0.10, 0.10}
\usepackage[pagebackref, breaklinks=true, colorlinks,citecolor=citecolor, linkcolor=linkcolor, bookmarks=false]{hyperref} % rgb
\usepackage{url}            % simple URL typesetting
\usepackage{booktabs}       % professional-quality tables
\usepackage{amssymb,amsmath,amsfonts}       % blackboard math symbols
\usepackage{nicefrac}       % compact symbols for 1/2, etc.
\usepackage{microtype}      % microtypography

\usepackage{float}%提供float浮动环境
\usepackage{booktabs}%提供命令\toprule、\midrule、\bottomrule
\usepackage{multirow}
\usepackage{makecell}
\usepackage{mathtools}
\usepackage{bbding}
\usepackage{subfigure}
\usepackage{enumitem}

\title{CyIN: Cyclic Informative Latent Space for Bridging Complete and Incomplete Multimodal Learning}

\author{Ronghao Lin$^{1,2}$,
~Qiaolin He$^1$,
~Sijie Mai$^3$,
~Ying Zeng$^1$,
~Aolin Xiong$^1$, \\
\textbf{~Li Huang$^{1,4}$, 
~Yap-peng Tan$^2$,
~Haifeng Hu$^{1,5}$}\thanks{~~Corresponding author.} 
\vspace{3pt}\\
$^1$ School of Electronics and Information Technology, Sun Yat-Sen University, China\\
$^2$ School of Electrical and Electronic Engineering, Nanyang Technological University, Singapore \\
$^3$ School of Computer Science, South China Normal University, China \\
$^4$ Desay SV Automotive Co., Ltd, China \\
$^5$ Pazhou Laboratory, China
\vspace{3pt}\\  
\texttt{\{linrh7,heqlin5,zengy268,xiongaolin\}@mail2.sysu.edu.cn,} \\ % dangjsh
\texttt{sijiemai@m.scnu.edu.cn,Li.Huang@desaysv.com}\\ 
\texttt{eyptan@ntu.edu.sg, huhaif@mail.sysu.edu.cn}
}

\begin{document}

\maketitle

\begin{abstract}
Multimodal machine learning, mimicking the human brain’s ability to integrate various modalities has seen rapid growth. Most previous multimodal models are trained on perfectly paired multimodal input to reach optimal performance. In real‑world deployments, however, the presence of modality is highly variable and unpredictable, causing the pre-trained models in suffering significant performance drops and fail to remain robust with dynamic missing modalities circumstances. In this paper, we present a novel Cyclic INformative Learning framework (CyIN) to bridge the gap between complete and incomplete multimodal learning. Specifically, we firstly build an informative latent space by adopting token- and label-level Information Bottleneck (IB) cyclically among various modalities. Capturing task-related features with variational approximation, the informative bottleneck latents are purified for more efficient cross-modal interaction and multimodal fusion. Moreover, to supplement the missing information caused by incomplete multimodal input, we propose cross-modal cyclic translation by reconstruct the missing modalities with the remained ones through forward and reverse propagation process. With the help of the extracted and reconstructed informative latents, CyIN succeeds in jointly optimizing complete and incomplete multimodal learning in one unified model. Extensive experiments on 4 multimodal datasets demonstrate the superior performance of our method in both complete and diverse incomplete scenarios. 
%Code is released at \url{https://github.com/RH-Lin/CyIN}. 
\footnote[1]{\ \ Code is released at \url{https://github.com/RH-Lin/CyIN}. }
%Code is released at \url{https://anonymous.4open.science/r/CyIN-1DD3}

\end{abstract}

% Keyword: Multimodal representation learning, Incomplete multimodal learning, Missing modality issue, Information bottleneck, Cross-modal transaltion

\section{Introduction}

To obtain the optimal multimodal performance, previous methods implicitly assume that every modality present at training will also be available at inference time. However, in real-world scenarios, multimodal data may be missing due to numerous factors such as fail sensors, causing uncertainty of the presence of input modalities \cite{wu2024deep}. The multimodal model exhibit pronounced sensitivity to such incomplete multimodal input, resulting in severe performance degradation when deploying the pre-trained multimodal model in downstream inference, especially for Transformer-based models \cite{ma2022multimodal}. This issue is summarized  as missing modality issue \cite{ma2021smil,lin2023missmodal,yuan2024noise}, and the methods devoted to addressing it are called incomplete multimodal learning methods \cite{li2024toward,lin2024adapt,zhang2024towards}.

Current methods focus on enhancing the robustness of multimodal models by designing various delicate modules in multimodal learning to deal with diverse missing circumstances, mainly divided into alignment and generation methods. The former leverage technologies such as contrastive learning \cite{poklukar2022geometric,lin2023missmodal,liu2024cifmmin}, canonical correlation analysis \cite{hotelling1992cca,andrew2013dcca,wang2015dccae}, and data augmentation like mixup or noisy input \cite{hazarika2022analyzing,lin2024adapt,yuan2024noise} to align the representations with complete and incomplete multimodal inputs, while the latter introduce generative models such as autoencoders \cite{bengio2006greedy,pham2019found,zhao2021missing}, variational autoencoders \cite{wu2018mvae,shi2019mmvae,vasco2022leveraging,vasco2022muse}, graph-based networks \cite{lian2023gcnet,shou2024adversarial}, and diffusion models \cite{wang2023incomplete} to reconstruct the missing information.

Although these methods have alleviated the missing modality issue by bridging the information gap in varying degrees, they suffer from insufficient exploitation in missing information and task-unrelated noise interruption no matter in alignment or generation \cite{wang2023incomplete,li2024toward}. Moreover, previous methods typically require training separate models tailored to each possible combination of missing modalities \cite{pham2019found,lian2023gcnet}. Consequently, their capacity in generalization and robustness maybe largely limited due to unknown and dynamic missing circumstances in real-world scenarios. Besides, most of previous methods inevitably sacrifice the complete multimodal performance when addressing incomplete input, failing in jointly combining complete and incomplete multimodal learning in a single model \cite{lin2024adapt}.

In this paper, we propose CyIN, a novel Cyclic INformative Learning framework that unifies complete and incomplete multimodal learning within a unified model. Firstly, we constructs an effective informative latent space via token- and label-level Information Bottlenecks (IB) to encourage the information flow in multimodal interaction. The former builds information bridge on token embeddings at low-level perception and adopt cyclic interaction among various unimodal representations, while the latter utilize the ground truth labels as guidance to introduce high-level semantics to the informative space. Combining these two IB objectives, we efficiently capture task-relevant features and filter out the redundant noise by sampling the bottleneck latents with variational approximation. Besides, to address missing modality issue, we introduce cross-modal cyclic translation to perform forward and reverse propagation between remained and missing modalities, thereby reconstructing the missing information in the informative space. By jointly optimizing both cyclic IB and translation 
process, CyIN seamlessly bridges complete and incomplete multimodal learning in a unified informative latent space. The key contributions of our paper can be summarized as:

\begin{itemize}
\item \textbf{Informative Latent Bottleneck Space.} Built by token- and label-level IB, the proposed informative latent space efficiently bridge complete and incomplete learning in one unified framework, where multimodal fusion and missing information reconstruction can both benefit from the purified bottleneck latents.

\item \textbf{Cyclic Interaction and Translation.} The presented cyclic information processing progress greatly boost the performance of cross-modal interaction in complete multimodal learning and enhance the translation quality in incomplete multimodal learning.

\item Extensive experiments on 4 datasets validate that CyIN achieves state-of-the-art performance in multimodal learning and maintain robustness across various missing modality scenarios.
\end{itemize}

\section{Preliminary}
\subsection{Complete and Incomplete Multimodal Representation Learning}
Considering multimodal inputs with multiple unimodal data source $X_u\in\{X_0, X_1,..., X_U\}$, multimodal learning aims at integrating complementary information from paired modalities $u\in\{u_0,u_1,...,u_U\}$ to learn multimodal representations that can drive inference on a variety of downstream tasks \cite{wang2023large,zhang2024mm}, where $|u|=U$ denotes the total number of modalities. The performance hinges on the strategy of multimodal fusion to effectively lverages informaiotn from each modality \cite{gkoumas2021makes,manzoor2023multmodality}, which in turn requires the completeness of modalities.  

\begin{figure}[t]
	\centering %居中
	\includegraphics[scale=0.55]{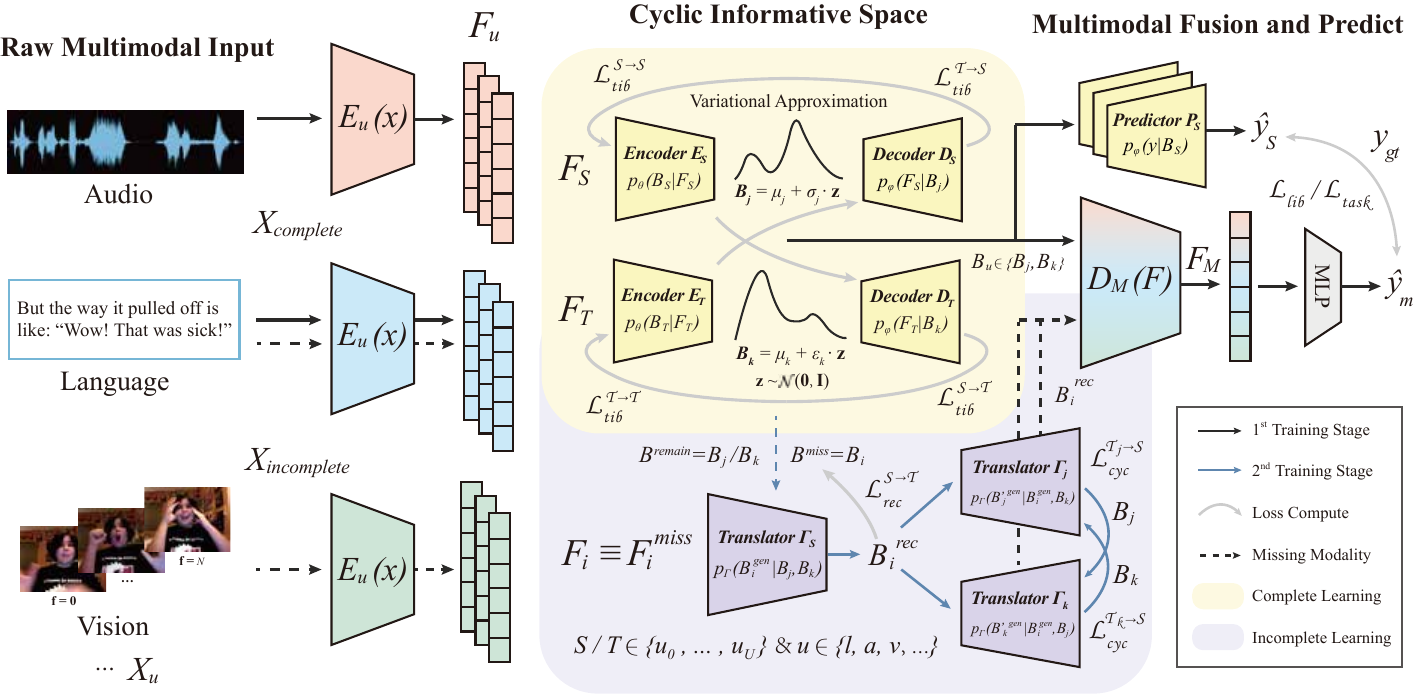}%图片大小及加载的图片名称	
	\caption{Framework overview. The proposed CyIN build a cyclic informative space to jointly train the complete and incomplete multimodal ealrning. }%图片标题
	\label{figure_framework}%标注该图片，用于在文章内引用
\end{figure}

As shown in Figure \ref{figure_framework}, each modality raw input is firstly processed by modality-specific encoders $E_u:X_u\mapsto F_u$ to produce unimodal representations $F_u$, and then merged by multimodal fusion decoders $D_M:\{F_u\}\mapsto F_M$ to generate the multimodal representations $F_M$. With multimodal input denoted as $X_u \equiv X_{complete}$, the process of complete multimodal learning can be formulated as:
\begin{equation}
    F_M=D_M(F_0,F_1,...,F_U),\ F_u=E_u(X_u),\ \text{where } u \in \{u_0,u_1,...,u_U\}
\label{equ_multimodal_fnn}
\end{equation}

When there are missing modalities in the multimodal inputs, denoted as $X_u \equiv X_{incomplete}=\{X^{remain}, X^{miss}\}$, the learning process of multimodal models is turned into incomplete multimodal learning, where $X^{remain}$ denotes the original unimodal data input while $X^{miss}$ denotes the zero vector without corresponding unimodal information. Thus, the incomplete multimodal representations $F_M$ in Equ. \ref{equ_multimodal_fnn} should be denoted as:
\begin{equation}
    F_M=D_M(F^{remain},F^{miss}),\ F^{remain}=E_u(X_u), F^{miss} = \textbf{0} \ \text{, where } u \in \{u_0,u_1,...,u_U\}
\end{equation}

The final prediction $\hat{y}_m$ is obtained by \textit{MLP} head on the multimodal representations $\hat{y}_M=MLP(F_M)$. Then, the optimization objective can be computed according to the specific regression (Mean Absolute Error) or classification (Cross Entropy) tasks, denoted as:
\begin{equation}
\mathcal{L}_{task} =\left\{
\begin{aligned}
    & \mathbb{E}\ |y_{gt} - \hat{y}_M| \ \text{, for regression }\\
    & -\mathbb{E}\ [y_{gt}\log\hat{y}_M]\ \text{, for classification }
    \end{aligned}
\right.\ \ 
\label{equ_task_loss}
\end{equation}
Due to the fact that the missing information will interrupt the original multimodal fusion space \cite{lian2023gcnet, lin2023missmodal} and semantic ambiguity issue may raised by overfitting on the remained modalities \cite{hazarika2022analyzing, ma2022multimodal}, the performance of incomplete multimodal learning decreases significantly compared with complete multimodal learning. Thus, the demand of jointly improving complete multimodal fusion and enhancing the reconstruction performance for incomplete input with missing modalities in one unified model remains challenging for the general multimodal systems \cite{lin2024adapt}.

\subsection{Information Bottleneck}
The approach of Information Bottleneck (IB) is proposed to compress the source state $S$ into a compact bottleneck latent $B$, which contains the  least necessary features from $S$ while preserves the most relevant information about the target state $T$ \cite{tishby1999information,tishby2015deepib,michael2018on}. Using mutual information to provide the the relevance between states, such trade-off can be written as the following minimization problem: 
\begin{equation}
\label{equ_mutual_info}
    \min_{p(B|S)} I(S;B) - \beta\ I(B;T)
\end{equation}
where $I(\cdot|\cdot)$ denotes the mutual information between two states and the hyper-parameter $\beta$ controls the trade-off degree. By introducing parameterized networks, the former mutual information term $I(B;S)$ can be modeled by IB encoder $E_S:S\mapsto B$ to extract information from source state, while the second term $I(T;B)$ can be conducted by IB decoder $D_T:B\mapsto T$ referring to the target state. 

For the purpose of optimizing, VIB \cite{alemi2017vib} utilize variational approximation to replace the intractable mutual information terms, yielding the following upper bound:
\begin{equation}
    I(S;B) - \beta\ I(B;T) \leq \mathbb{E}_{S\sim p(S)} \left[KL( p_\theta(B|S)\parallel q(B))\right] - \beta\ \mathbb{E}_{B\sim p(B|S)}\mathbb{E}_{S\sim p(S)} [\log q_\phi(T|B)]
\label{equ_vib}
\end{equation}
where $p_\theta(B|S)\sim\mathcal{N}(\mu, \sigma^2)$ and $q(B)\sim\mathcal{N}(0,\textbf{I})$ satisfy the Gaussian Distribution. Denoting loss objective $\mathcal{L}_{ib}=I(B;S) - \beta\ I(T;B)$, optimizing $\mathcal{L}_{ib}$ is equivalent to minimizing the upper bound in Equ. \ref{equ_vib}. Details derivations are presented in Appendix \ref{appendix_vib}. 

Utilizing reparameterization trick \cite{kingma2014autoencoding,alemi2017vib} to encourage gradient propagation, the bottleneck latents $B$ can be sampled from $p_\theta(B|S)$, denoted as:
\begin{equation}
    B=\mu+\sigma\ \odot\ \textbf{z} \ \text{, where } \textbf{z} \sim \mathcal{N}(0,\textbf{I})
\label{equ_reparameterization}
\end{equation}
Serving as a bridge carrying compactly relevant information between $S$ and $T$, the bottleneck latent $B$ can effectively construct an informative space to control the information flow from the source state $S$ to the target state $T$. More related works and background can be found in Appendix \ref{appendix_relatedwork}

% \textbf{Unconditinal and Conditional generation}

% \textbf{Cross-modal generation}

\section{Methodology}
\subsection{Multimodal Informative Latent Space}
% Without losing versatility, 
As shown in Figure \ref{figure_framework} , both complete and incomplete multimodal learning require task-relevant informative representations, regardless of modeling intra- or inter-modal relationships. Therefore, constructing a unified informative latent space can not only be beneficial in bridging the modality gap in the fusion of multimodal representations, but strengthening the reconstruction of the most meaningful features among various modalities. To enhance the efficiency of informative space, we introduce two complementary IB mechanisms to encourage the compactness of the bottleneck latents at both perceptual low-level and semantics high-level, referring to token- and label-level IB.

\textbf{Token-level Information Bottleneck.} Without losing generality, for arbitrary two modalities in multimodal inputs $X_u\in\{X_0,...,X_U\}$, we denote one modality input $X_S$ as the source state while another one $X_T$ as the target state in Equ. \ref{equ_mutual_info}. Let $X_S/X_T=\{x_u^0, ... ,x_u^{L}\}$ denote the sequence of token-level inputs for modality $u$ where $S/T\in\{u_0,u_1,..,u_U\}$, where $L$ denotes the sequence length. Then, producing by the corresponding modality-specific encoder $E_u:X_u\mapsto F_u$, the unimodal representations can be obtained as $F_S/F_T=\{f_u^i\}|_{i=1}^{L}\in\mathbb{R}^{L\times C}$, where $f_u$ denotes token embedding and $C$ denotes the feature dimension of specific modality.

With IB encoder $E_S:F_S\mapsto B_S$, the token-level bottleneck latents $B_S=\{b_S^i\}|_{i=1}^{L}$ can be attained by applying information bottleneck on all token embeddings, denoted as:
\begin{equation}
\begin{aligned}
    \mathcal{L}^{S\rightarrow T}_{tib} 
    &\approx \frac{1}{L}\sum^L_{i} \{\mathbb{E}_{f^i_S\sim p(f^i_S)} \left[KL( p_\theta(b^i_S|f^i_S)\parallel q(b^i_S))\right] - \beta\ \mathbb{E}_{b^i_S\sim p(b^i_S|f^i_S)}\mathbb{E}_{f^i_S\sim p(f^i_S)} [\log q_\phi(f^i_T|b^i_S)] \}
    % \mathcal{L}^{j\rightarrow k}_{ib} &\approx \mathbb{E}_{F_j\sim p(F_j)} KL( p_\theta(B_j|F_j)\parallel q(B_j)) - \beta\ \mathbb{E}_{B_j\sim p(B_j|F_j)}\mathbb{E}_{F_j\sim p(F_j)} [\log q_\phi(F_k|B_j)]\\
    % &= \frac{1}{L}\sum^L_{i} \{\mathbb{E}_{f^i_j\sim p(f^i_j)} KL( p_\theta(b^i_j|f^i_j)\parallel q(b_j)) - \beta\ \mathbb{E}_{b^i_j\sim p(b^i_j|f^i_j)}\mathbb{E}_{f^i_j\sim p(f^i_j)} [\log q_\phi(f^i_k|b^i_j)] \}
\label{equ_token_ib1}
\end{aligned}
\end{equation}

% we can formulate posterior probability $q_\phi(f_T|b_S)=e^{-\parallel f_T-D_T(b_S)\parallel^2}$ with IB decoder $D_T:B_S\mapsto F_T$.
Since unimodal representations $F_S/F_T$ are concatenated by token embeddings with continuous value, we can formulate posterior probability $q_\phi(f_T|b_S)$ with IB decoder $D_T:B_S\mapsto F_T$ to project bottleneck of the source representation into the target one. Besides, given $p_\theta(b^i_S|f^i_S)\sim\mathcal{N}(\mu^i_B, (\sigma_B^{i})^2)$ and $q(b^i_S)\sim\mathcal{N}(0,\textbf{I})$, then Equ. \ref{equ_token_ib1} can be derived as
\begin{equation}
    \mathcal{L}^{S\rightarrow T}_{tib} \approx \frac{1}{L}\sum^L_{i} \{KL( \mathcal{N}(\mu^i_B, (\sigma_B^{i})^2)\parallel\mathcal{N}(0,\textbf{I})) + \beta\ \mathbb{E}_{b_S} [\parallel f_T-D_T(b_S)\parallel^2] \}
\label{equ_token_ib2}
\end{equation}

\textbf{Cyclic Interaction.} With unimodal representations $F_u\in\{F_0,...,F_U\}$ from various modalities, the source and target modalities in token-level IB can be cyclically chosen in an iterative way to filter the redundant noise contained in each unimodal representation and enhance cross-modal interaction. 

Considering the source state and target state as the same modality when $F_S=F_T$, the optimization of token-level IB focuses on learning modality-specific features and capturing intra-modal dynamics, denoted as $\mathcal{L}^{S\rightarrow S}_{tib}$. The two terms in Equ. \ref{equ_token_ib2} focuses on compressing sufficient information from each token embedding and re-projecting to the original sequence, respectively. 

On the other hand, when the source state and target state comes from diverse modalities with $F_S\neq F_T$, the token-level IB aims at modeling invariant information across different modalities and seizing the inter-modal dynamics. Then Equ. \ref{equ_token_ib2} aims at integrating modality-shared features for the source modality embeddings to match the target one. Besides, interchanging the role of modalities as source and target states in Equ. \ref{equ_token_ib2}, referring as $\mathcal{L}^{S\rightarrow T}_{tib}$ and $\mathcal{L}^{T\rightarrow S}_{tib}$, the modality-shared features can be further explored by such directional information flow.

% Since the sequence length maybe various for different modalities, we utilize an alignment network with Conv1D average pooling to automatically align the token embeddings for each modality. 

Combing the intra- and inter-modal settings, the final loss of token-level IB can be denoted as:
\begin{equation}
    \mathcal{L}_{tib} =  \mathbb{E}_{S\cup T}[\mathcal{L}^{S\rightarrow S}_{tib} + \frac{1}{2}(\mathcal{L}^{S\rightarrow T}_{tib} + \mathcal{L}^{T\rightarrow S}_{tib})] \ \text{, where } S/T\in\{u_0,...,u_U\} 
\label{equ_token_ib_final}
\end{equation}

\textbf{Label-level Information Bottleneck.} Except for token-level IB focusing on low-level perception, we introduce label-level IB to inject the supervision of high-level semantics in the information flow. For unimodal representation $F_S$ from each modality $X_S$ as the source state, we utilize the ground truth label $y_{gt}$ (regression scores or recognition classes) as the supervised target. Given $N$ batched samples $\{X^i_S\}|_{i=1}^{N}$ for each modality, with IB encoder $E_S:F_S\mapsto B_S$ and predictor $P_S:B_S\mapsto \hat{y}_S$, the label-level bottleneck latents can be constrained with the following loss:
\begin{equation}
    \small{\mathcal{L}^{S}_{lib} \approx \frac{1}{N}\sum^N_{i} \{\mathbb{E}_{F^i_S\sim p(F^i_S)}\left[ KL( p_\theta(B^i_S|F^i_S)\parallel q(B_S))\right] - \beta\ \mathbb{E}_{B^i_S\sim p(B^i_S|F^i_S)}\mathbb{E}_{F^i_S\sim p(F^i_S)} [\log q_\phi(y_{gt}|B^i_S)] \}}
\label{equ_label_ib1}
\end{equation}

Iterating computing $\mathcal{L}^{S}_{lib}$ through all input modalities $S\in\{u_0,u_1,...,u_U\}$, we can derive the overall label-level IB loss. In practice, for regression task, the posterior probability $q_\phi(y_{gt}|B_S)$ is formulated as $q_\phi(y_{gt}|B_S)=e^{-\parallel y_{gt}-\hat{y}_S\parallel}=e^{-\parallel y_{gt}-P_S(B_S)\parallel}$, then we have:
\begin{equation}
    \mathcal{L}_{lib} \approx \mathbb{E}_{S}\{\frac{1}{N}\sum^N_{i} \left[KL( \mathcal{N}(\mu^i_B, (\sigma_B^{i})^2)\parallel\mathcal{N}(0,\textbf{I}))\right] + \beta\ \mathbb{E}_{B_S} [\parallel y_{gt}-P_S(B_S)\parallel]\}
\label{equ_label_ib_final1}
\end{equation}

%(1-P_S(B_S))^{1-y_{gt}}
While for classification task with $V$ classes, the posterior probability $q_\phi(y_{gt}|B_S)$ is computed as $q_\phi(y_{gt}|B_S)=\prod^V \hat{y}_S^{y_{gt}}=\prod ^V[P_S(B_S)]^{y_{gt}}$, then we have:
\begin{equation}
    \mathcal{L}_{lib} \approx \mathbb{E}_{S}\{\frac{1}{N}\sum^N_{i} \left[KL( \mathcal{N}(\mu^i_B, (\sigma_B^{i})^2)\parallel\mathcal{N}(0,\textbf{I}))\right] - \beta\ \mathbb{E}_{B_S} [\sum^V y_{gt}\log{P_S(B_S)}]\}
\label{equ_label_ib_final2}
\end{equation}

Optimized by token- and label-level IB loss, we can build an informative space which controls the information flow inside and across modalities in the guidance of task-related semantics.
\subsection{Cross-modal Cyclic Translation}
Since the bottleneck latents are learned in the informative space, task-irrelevant features and unimodal inherent noise are sufficiently filtered out. Hence, with incomplete multimodal inputs, reconstructing missing information in the built informative latent space becomes significantly easier than attempting reconstruction in the original space. %The ablation study and further analysis in Section \ref{experiments} supports this conclusion, highlighting improvements in both performance and convergence stability. 
Here we present the cross-modal cyclic translation with forward and reverse propagation to enhance model's robustness under incomplete multimodal scenarios.

\textbf{Forward Propagation.} In order to reconstruct the missing information, we leverage Cascaded Residual Autoencoder (CRA) \cite{tran2017cra} as the translator $\Gamma_{S\rightarrow T}:B_S\mapsto B_T$ across various modalities as the machine translation task \cite{bahdanau2015neural}. CRA has been demonstrated sufficiently in mitigating the modality gap and translating information from the source modality $S$ to the target one $T$, denoted as:
\begin{equation}
    B_{S\rightarrow T}^{rec} =\Gamma_{S\rightarrow T}(B_S) = p_\Phi(B_T|B_S)
\end{equation}
where translator $\Gamma$ consists of a series of stacked Residual Autoencoders $RA^S_i(\cdot)$ denoted as:
\begin{equation}
    r_T=\left\{
    \begin{aligned}
    & RA^S_1(B_S),\ i=1 \\
    & RA^S_i(B_S+ \sum\nolimits^{n-1}_{i=1} RA^S_i(B_S)),\ i>1
    \end{aligned}
    \right.
\end{equation}
where $r_T$ denotes the output of each $RA$ block and the last output of $RA_i$ is the overall translated information denoted as $B_{S\rightarrow T}^{rec}$. For the translation process $S\rightarrow T$, we align the translated information with the original one by a reconstruction loss computed as:
\begin{equation}
    \mathcal{L}^{S\rightarrow T}_{rec} =\parallel B_T - B_{S\rightarrow T}^{rec} \parallel^2
\end{equation}

\textbf{Reverse Propagation.} To improve translation performance and encourage sufficient exploration of inter-modal dynamics, we further apply cyclic consistent learning \cite{wang2020transmodality,zhao2021missing} to reverse the translated direction of information flow as back-translation trick \cite{hoang2018iteractivebt}. Since forward propagation denotes the reconstruction process of information flow $B_S\rightarrow B^{rec}_{S\rightarrow T}$ from source to target modality, we adopt reverse propagation to translate the reconstructed information back as $B^{rec}_{S\rightarrow T} \rightarrow B^{cyc}_S$, denoted as:
\begin{equation}
    B_S^{cyc} =\Gamma_{T\rightarrow S}(B_{S\rightarrow T}^{rec}) = p_\Phi(B_S|B_{S\rightarrow T}^{rec})
\end{equation}
where translator $\Gamma_{T\rightarrow S}$ shares the model weights with the one used in forward propagation. Thus, the reconstruction loss for the reverse propagation can be denoted as:
\begin{equation}
    \mathcal{L}^{T\rightarrow S}_{cyc} =\parallel B_S - B_S^{cyc} \parallel^2
\end{equation}

\textbf{Generalize to Multiple Remained Modalities.} For multimodal learning with more than two modalities as input, the missing circumstance of modalities is highly uncertain when encountering incomplete multimodal input. When the number of remained modalities are more than one, how to effectively integrating features from these modalities to reconstruct the missing one is yet to address due to the diverse incomplete input circumstances. Thus, we present a generalizable solution to employ the paired cross-modal translator to generate missing information regardless of the number of remained modalities.

Considering data $X^{incomplete}\equiv \{X^{remain}, X^{miss}\}$ with remained $\{u_j,...,u_k\}$ and missing $u_i$ modality, we aims at translating the informative bottleneck latents $B^{remain}=\{B_j,...,B_k\}$ of the remained modalities $X^{remain}$ into the latents $B^{miss}$ of missing modality $X^{miss}$, denoted as:
\begin{equation}
    B^{miss}\coloneqq B_{i}^{rec} =\Gamma_u(B^{remain}) = p_\Phi(B_i|B_j,...,B_k)
\end{equation}

Instead of training more translators to fit in various circumstances of the remained modalities input, we leverage the additive characteristic of Gaussian Distribution to integrate the translated missing information. Since the original bottleneck latents $\{B^{remain}, B^{miss}\}$ are all sampled from the informative space, we directly combine the outputs of each  cross-modal translator according to the source modality and denote the translated latents of missing modalities as:
\begin{equation}
    B_{i}^{rec}=\sum^{|u|}_{j\neq i} B_{j\rightarrow i}^{rec}=\sum^{|u|}_{j\neq i} \Gamma_{j\rightarrow i}(B_j) = \prod^{|u|}_{j\neq i} p_\Phi^j(B_i|B_j)
\end{equation}
where $B_{i}^{rec}$ can be denoted as the translation from a special form of Gaussian Mixture Model $\sum \Gamma_{j\rightarrow i}(\mathcal{N}(\mu_j,\sigma^2_j))$. Regardless of the specific settings of remained modalities, we utilize $B_{i}^{rec}$ as the final translated informative latents to supplement the information of missing modality $B^{miss}$.

% Aligning $B^{rec}_i$ with $B_i$

Combining forward and reverse propagation, the objective for cross-modal cyclic translation is:
\begin{equation}
    \mathcal{L}_{tran} = \mathcal{L}^{S\rightarrow T}_{rec} + \mathcal{L}^{T\rightarrow S}_{cyc}\ \text{, where } S/T\in\{u_i,u_j,...,u_k\} \text{ and } S\neq T
\label{equ_translation}
\end{equation}

\subsection{Complete and Incomplete Multimodal Fusion}
We jointly conduct complete and incomplete multimodal learning in one unified multimodal fusion network, boosting the robustness with incomplete input and maintaining the performance with complete ones. For simplicity, we employ Multimodal Transformer \cite{tsai2019multimodal} to present multimodal fusion decoder $D_M:\{B_0,B_1,...,B_U\}\mapsto F_M$ for bottleneck latents output by the cyclic information space. Note that the fusion module can be replaced by any networks.

Considering bottleneck latents of arbitrary two modalities $\{B_j,B_k\}$, the multimodal fusion ecoder $D_M$ consists of a series of multi-head cross-modal attention layers, composed of:
\begin{equation}
\begin{aligned}
&y_r^h=CMAttention_r(B_j,B_k)=Softmax\left[B_jW^h_q\cdot (W^h_kB_k)^T/\sqrt{C} \right]B_kW^h_v \\
&Y_r^H=Concat(y_r^1,..,y_r^h)W^o \\
&Z_{r}=Y^H_r+LayerNorm(Y^H_{r-1})\ \text{, where } r\in\{1,...,R\} \\
&M_{r}=FeedForward(LayerNorm(Z_{r}))+LayerNorm(Z_{r})
\end{aligned}
\end{equation}
where $H$ denotes the number of heads, $C$ denotes the dimension of bottleneck latents, $W_q,W_k,W_v,W_o$ denotes the weight matrices and $R$ denotes the number of cross-modal attention layers. The output of last layer $M^{j,k}_{R}$ is used as the bimodal fusion latent of modalities $\{u_j,u_k\}$. 

Iteratively extracting $M^{j,k}_{R}$ with paired modalities from $u\in\{u_0,...,u_U\}$ for both complete and incomplete multimodal learning, the final multimodal representation can be denoted as:
\begin{equation}
F_M=\left\{
\begin{aligned}
&D_M(B_0,B_1,...,B_U)\\
&D_M(B^{remain},B^{miss})
\end{aligned}
\right\}=Concat([M^{0,1}_{R},M^{1,2}_{R},...,M^{U-1,U}_{R}])
\end{equation}
\textbf{Optimization Objective} To sum up, our framework involves four learning objectives, including task prediction loss $\mathcal{L}_{task}$ in Equ. \ref{equ_task_loss}, token-level information bottleneck loss $\mathcal{L}_{tib}$ in Equ. \ref{equ_token_ib_final}, label-level information bottleneck loss $\mathcal{L}_{lib}$ in Equ. \ref{equ_label_ib_final1} or Equ. \ref{equ_label_ib_final2}, and cross-modal cyclic translation loss $\mathcal{L}_{tran}$ in Equ. \ref{equ_translation}. The total loss can be written as:
\begin{equation}
\mathcal{L}_{total}=\mathcal{L}_{task}+\frac{1}{\beta}(\mathcal{L}_{tib}+\mathcal{L}_{lib})+\gamma \mathcal{L}_{tran}
\label{equ_loss_total}
\end{equation}
where $\beta$ denotes the balancing trade-off weight of mutual information among bottleneck latents and representations and $\gamma$ denotes the weight of translation training for incomplete multimodal learning. % both $\beta$ and $\gamma$ are hyper-parameters.

\textbf{Multi-stage Training} 
Since both complete and incomplete multimodal learning requires an effective informative bottleneck space, we divide the training process into two stages. The first stage set $\gamma=0$ to make the multimodal learning focus on the construction and stabilization of informative space under complete multimodal learning, while the second stage set $\gamma>0$ to gradually introduce the training of cross-modal translator to enhance the incomplete multimodal learning. 

\section{Experiments}
\label{experiments}
In this section, we conduct comprehensive experiments for complete and incomplete multimodal learning on 4 datasets to evaluate the performance of the proposed framework.

\textbf{Tasks and Datasets.} \textit{Multimodal Regression task}: \textbf{MOSI} and \textbf{MOSEI} are multimodal sentiment analysis datasets contain 2,199 and 22,856 YouTube opinion video clips, respectively, where each clip is annotated with a continuous sentiment score ranging from –3 (strongly negative) to +3 (strongly positive) as Likert scale. \textit{Multimodal Classification task}: \textbf{IEMOCAP} dataset provides 7,369 video-recoded conversation, annotated with six emotion categories: $\{$happy, sad, neutral, angry, excited, and frustrated$\}$.  \textbf{MELD} dataset includes 13,391 utterances drawn from multi-party conversations of television series Friend, labeled with seven emotions: $\{$neutral, surprise, fear, sadness, joy, disgust, and anger$\}$. These two datasets are provided for multimodal emotion recognition.

\begin{table}[htbp]
\centering
\setlength\tabcolsep{7.5pt}
\caption{Performance comparison between the proposed CyIN and baselines with the average results in complete modality setting $u\in\{l,a,v\}$ and incomplete modality settings, with fixed missing protocols including modality settings $u\in\{l\}/\{v\}/\{a\}/\{l,a\}/\{l,v\}/\{a,v\}$ and random missing protocols including missing rates $MR\in[0.1,0.2,0.3,0.4,0.5,0.6,0.7]$}
\label{table_all}
% \resizebox{0.8\textwidth}{!}{%
% \begin{adjustbox}{scale=0.65}
\scalebox{0.65}{
\begin{tabular}{cccccccccccccc}
    \toprule[1.5pt]
    \multirow{2}{*}{Setting} & \multirow{2}{*}{Models} & \multicolumn{4}{c}{MOSI} & \multicolumn{4}{c}{MOSEI} & \multicolumn{2}{c}{IEMOCAP} & \multicolumn{2}{c}{MELD}\\
    \cmidrule(r){3-6}\cmidrule(r){7-10}\cmidrule(r){11-12}\cmidrule(r){13-14}
    & & Acc7$\uparrow$ & F1$\uparrow$ & MAE$\downarrow$ & Corr$\uparrow$ & Acc7$\uparrow$ & F1$\uparrow$ & MAE$\downarrow$ & Corr$\uparrow$ & Acc$\uparrow$ & wF1$\uparrow$ & Acc$\uparrow$ & wF1$\uparrow$\\
    \midrule[1.5pt]
    \multirow{11}{*}{\makecell[c]{Complete\\$u\in\{l,a,v\}$}}  ~ & CCA & 27.7  & 74.9  & 1.106  & 0.541  & 46.1  & 82.9  & 0.654  & 0.666  & 61.7  & 61.5  & 51.2  & 46.6  \\ 
        ~ & DCCA & 25.1  & 73.6  & 1.220  & 0.422  & 39.3  & 73.3  & 0.787  & 0.425  & 56.8  & 55.5  & 47.7  & 37.0  \\ 
        ~ & DCCAE & 19.7  & 69.5  & 1.642  & 0.357  & 38.9  & 73.5  & 0.782  & 0.437  & 57.4  & 56.4  & 48.0  & 36.9  \\ 
        ~ & CPM-Net & 16.4  & 65.5  & 1.337  & 0.348  & 35.9  & 75.4  & 0.873  & 0.375  & 55.7  & 56.2  & 42.3  & 38.0  \\ 
        ~ & CRA & 34.8  & 83.2  & 0.916  & 0.741  & 51.4  & 85.5  & 0.553  & 0.765  & 63.4  & 62.2  & 57.6  & 54.8  \\ 
        ~ & MCTN & 43.0  & 84.6  & 0.752  & 0.783  & 47.9  & 84.2  & 0.592  & 0.721  & 58.4  & 57.8  & 56.3  & 52.4  \\ 
        ~ & MMIN & 43.2  & 85.0  & 0.744  & 0.782  & 52.9  & 84.9  & 0.537  & 0.769  & 62.5  & 62.7  & 60.6  & 56.2 \\ 
         ~ & GCNet & 43.6  & 85.8  & 0.732  & 0.792  & 52.6  & 85.9  & 0.531  & 0.778  & 63.0  & 63.0  & 60.4  & 58.5  \\ 
        ~ & IMDer & 43.8  & 85.7  & 0.724  & 0.796  & \textbf{53.8}  & 85.1  & 0.532  & 0.756  & 64.4  & 64.8  & 61.1  & 59.7 \\
        ~ & LNLN & 44.0  & 84.3  & 0.762  & 0.766  & 52.6  & 85.1  & 0.542  & 0.772  & 62.9  & 62.5  & 58.2  & 57.1  \\ 
    ~ & \textbf{CyIN}  & \textbf{48.0} & \textbf{86.3} & \textbf{0.712} & \textbf{0.801} & 53.2 & \textbf{86.1} & \textbf{0.530} & \textbf{0.774} & \textbf{66.1} & \textbf{66.0} & \textbf{61.6} & \textbf{59.8} \\ 
    % ~ & \textbf{CyIN}  & \textbf{ } & \textbf{ } & \textbf{ } & \textbf{ } & \textbf{ } & \textbf{ } & \textbf{ } & \textbf{ } & \textbf{ } & \textbf{ } & \textbf{ } & \textbf{ } \\ 
    \midrule[0.5pt]
    \multirow{11}{*}{\makecell[c]{Fixed\\Missing}}
    ~ & CCA & 21.8  & 57.7  & 1.264  & 0.339  & 43.1  & 69.7  & 0.744  & 0.446  & 45.9  & 41.0  & 49.1  & 39.4   \\ 
        ~ & DCCA & 19.7  & 58.8  & 1.418  & 0.261  & 41.3  & 70.0  & 0.799  & 0.392  & 41.3  & 38.9  & 47.0  & 34.4   \\ 
        ~ & DCCAE & 21.6  & 62.8  & 1.444  & 0.306  & 39.5  & 67.6  & 0.806  & 0.370  & 42.2  & 40.2  & 46.9  & 33.9   \\ 
        ~ & CPM-Net & 17.1  & 60.1  & 1.353  & 0.332  & 38.6  & 73.8  & 1.095  & 0.139  & 41.3  & 39.8  & 33.8  & 32.8   \\ 
        ~ & CRA & 27.3  & 67.8  & 1.158  & 0.404  & 45.4  & 78.5  & 0.672  & 0.593  & 47.9  & 45.9  & 50.7  & 45.4   \\ 
        ~ & MCTN & 28.5  & 63.0  & 1.104  & 0.392  & 44.9  & 73.2  & 0.717  & 0.409  & 40.6  & 34.3  & 52.4  & 42.0   \\ 
        ~ & MMIN & 31.3  & 68.4  & 1.093  & 0.433  & 46.2  & 77.7  & 0.661  & 0.588  & 50.8  & 50.2  & 53.9  & 43.4  \\ 
        ~ & GCNet & 29.5  & 69.5  & 1.065  & 0.538  & 45.5  & 73.6  & 0.697  & 0.551  & 52.8  & 51.9  & 50.4  & 46.7   \\ 
        ~ & IMDer & 31.4  & 70.6  & 1.043  & 0.533  & 47.1  & 76.6  & 0.680  & 0.583  & 54.7  & 54.4  & 53.1  & \textbf{49.8}  \\ 
        ~ & LNLN & 29.7  & 64.8  & 1.102  & 0.428  & 46.9  & 77.6  & 0.663  & 0.581  & 53.6  & 52.1  & 49.6  & 44.7  \\ 
    ~ & \textbf{CyIN}  & \textbf{32.8} & \textbf{72.2} & \textbf{1.037} & \textbf{0.599} & \textbf{47.6} & \textbf{78.6} & \textbf{0.656} & \textbf{0.594} & \textbf{57.4} & \textbf{56.6} & \textbf{54.4} & 49.4 \\ 
    \midrule[0.5pt]
    \multirow{11}{*}{\makecell[c]{Random\\Missing}} 
     ~ & CCA & 23.1  & 66.3  & 1.220  & 0.420  & 44.1  & 74.8  & 0.725  & 0.526  & 49.9  & 49.3  & 48.8  & 39.8   \\ 
        ~ & DCCA & 21.9  & 64.8  & 1.279  & 0.323  & 40.9  & 68.6  & 0.797  & 0.380  & 43.3  & 42.2  & 47.3  & 33.8   \\ 
        ~ & DCCAE & 19.7  & 62.7  & 1.566  & 0.276  & 39.0  & 65.4  & 0.810  & 0.345  & 43.2  & 41.7  & 47.1  & 34.3   \\ 
        ~ & CPM-Net & 16.9  & 63.9  & 1.340  & 0.325  & 34.3  & 74.0  & 1.497  & 0.078  & 53.1  & 53.7  & 41.6  & 33.6   \\ 
        ~ & CRA & 28.6  & 68.4  & 1.145  & 0.558  & 46.6  & 80.1  & 0.647  & 0.635  & 49.9  & 49.4  & 52.0  & 47.6   \\ 
        ~ & MCTN & 30.4  & 67.2  & 1.052  & 0.573  & 45.4  & 73.2  & 0.692  & 0.550  & 35.3  & 37.5  & 52.4  & 44.1   \\ 
        ~ & MMIN & 33.3  & 70.9  & 1.014  & 0.584  & 47.5  & 79.3  & 0.644  & 0.635  & 49.7  & 49.7  & 53.9  & 45.8  \\ 
        ~ & GCNet & 33.8  & 73.8  & 0.989  & 0.623  & 46.6  & 79.2  & 0.680  & 0.630  & 55.3  & 55.3  & 47.8  & 47.7   \\ 
        ~ & IMDer & 34.6  & 74.9  & 0.950  & 0.644  & 47.3  & 78.9  & 0.660  & 0.611  & 55.8  & 56.1  & 52.1  & 49.5  \\ 
        ~ & LNLN & 34.2  & 72.8  & 0.978  & 0.627  & 47.9  & 79.2  & 0.639  & 0.642  & 55.5  & 55.8  & 51.7  & 49.0  \\ 
    ~ & \textbf{CyIN}  & \textbf{35.0} & \textbf{75.7} & \textbf{0.943} & \textbf{0.650} & \textbf{48.3} & \textbf{79.9} & \textbf{0.633} & \textbf{0.650} & \textbf{57.5} & \textbf{57.5} & \textbf{54.8} & \textbf{50.5} \\ 
    \bottomrule[1.5pt]
\end{tabular}
}
% \end{adjustbox}
% \vspace{-1cm}
\end{table}

\textbf{Implementation Details.}
All experiments are performed on H800 GPU with Pytorch 2.4.1 on CUDA 12.4. For audio and vision modality, we leverage ImageBind \cite{girdhar2023imagebind} as a feature extractor for better alignment performance. For language modality, we use pre-trained BERT \cite{devlin2019bert} on MOSI and MOSEI while sBERT \cite{reimers2019sbert} on IEMOCAP and MELD for fair comparison with baselines. Detailed about hyper-parameters in each dataset is presented in Appendix \ref{appendix_hyperparam}.

\textbf{Evaluation Protocols.} Following \cite{zhao2021missing,lian2023gcnet, wang2023incomplete,zhang2024towards}, we evaluate the performance of different methods by (1) complete multimodal input ($u\in\{l,a,v\}$) and under (2) fixed missing protocol ($u\in\{l\}/\{v\}/\{a\}/\{l,a\}/\{l,v\}/\{a,v\}$) and (3) random missing protocol with missing rates $MR\in[0.1,0.2,0.3,0.4,0.5,0.6,0.7]$. Here $u\in\{l,a,v\}$ denote language, audio and vision modalities. Note that diverse with previous methods, we leverage one unified model for evaluation on various input circumstances with 10 runs. The practical details of missing protocols can be found in Appendix \ref{appendix_miss_protocol}. Details about baselines and evaluation metrics can be found in Appendix \ref{appendix_baseline}-\ref{appendix_metric}.

\textbf{Quantitative Comparisons.} The average results under both complete input or various incomplete multimodal input circumstances are reported in Table \ref{table_all}. Compared with previous baselines, CyIN reaches superior performance on most metrics in scenarios with both complete and incomplete multimodal input. 
Specifically, with the purify capability of informative space, CyIN sufficiently explore the modality-specific and -shared features and conduct efficient multimodal fusion based on the bottleneck latents when all multimodal input are present.

Under fixed missing protocols, CyIN consistently outperforms baselines across fixed one or two missing modality settings, demonstrating strong generalization even without specialized tuning for any specific modalities. The ability to integrate informatio from both dominant and inferior modalities highlights the flexibility of the informative bottleneck space. Besides under random missing protocols, CyIN remarkably enhances the model's robustness to various random modality missing scenarios, with minimal performance loss even at severe missing rate. The result illustrates broader application of CyIN for downstream multimodal tasks as in real-world where the presence of modalities is highly dynamic and unpredictable. More results and analysis are reported in Appendix \ref{appendix_more_result}

\begin{figure*}[htbp]
% \vspace{-0.3cm}
\centering
 \hspace{-1.5cm}\subfigure[Feature Distribution]{
 \begin{minipage}[bthp]{0.33\linewidth}
    % \hspace{-1cm}
    \centering
    \includegraphics[scale=0.12]{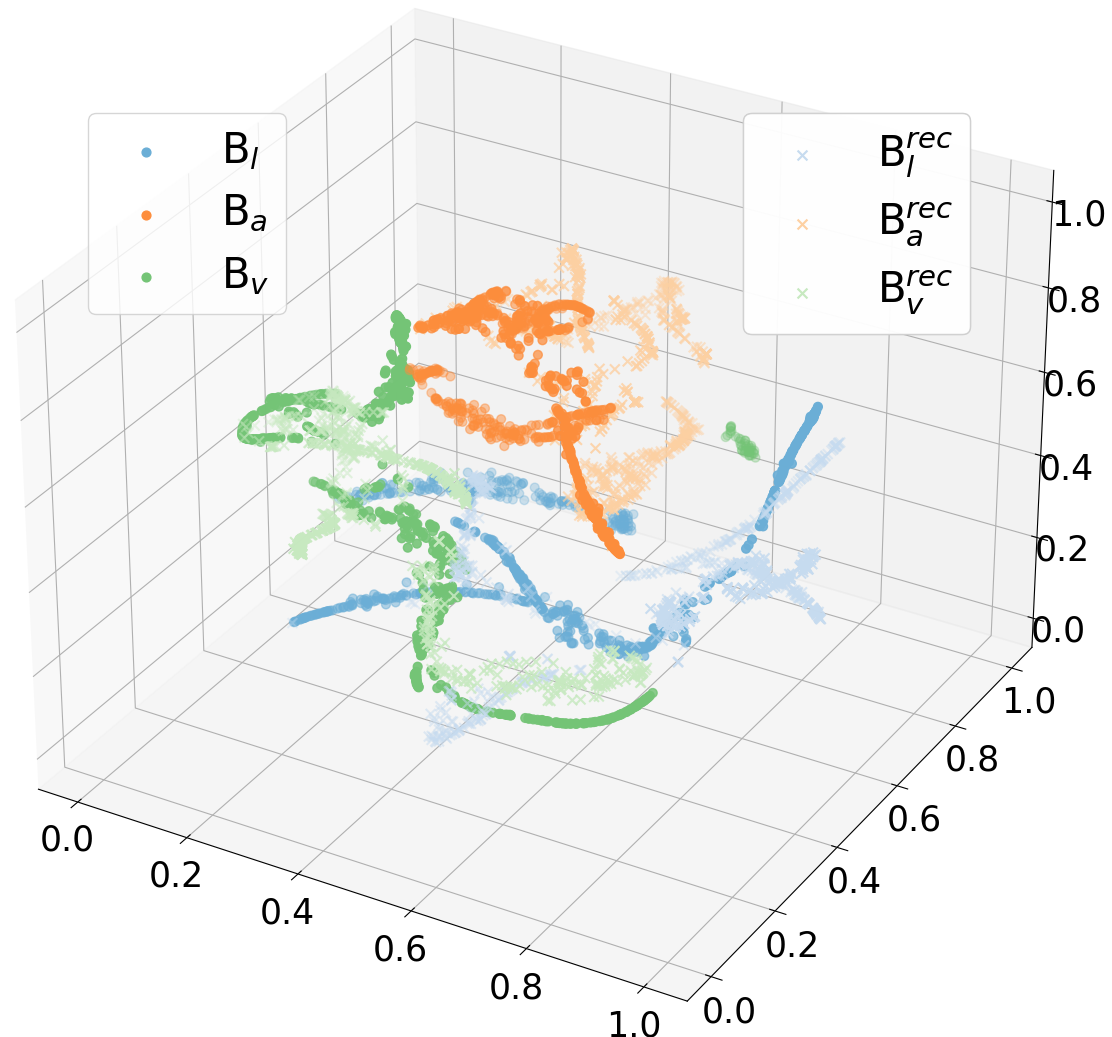} 
    % {figs/origin_transalte_latent_MOSI.pdf} 
    % \vspace{0.1cm}
    \\
    % \hspace{-1cm}
    \includegraphics[scale=0.12] {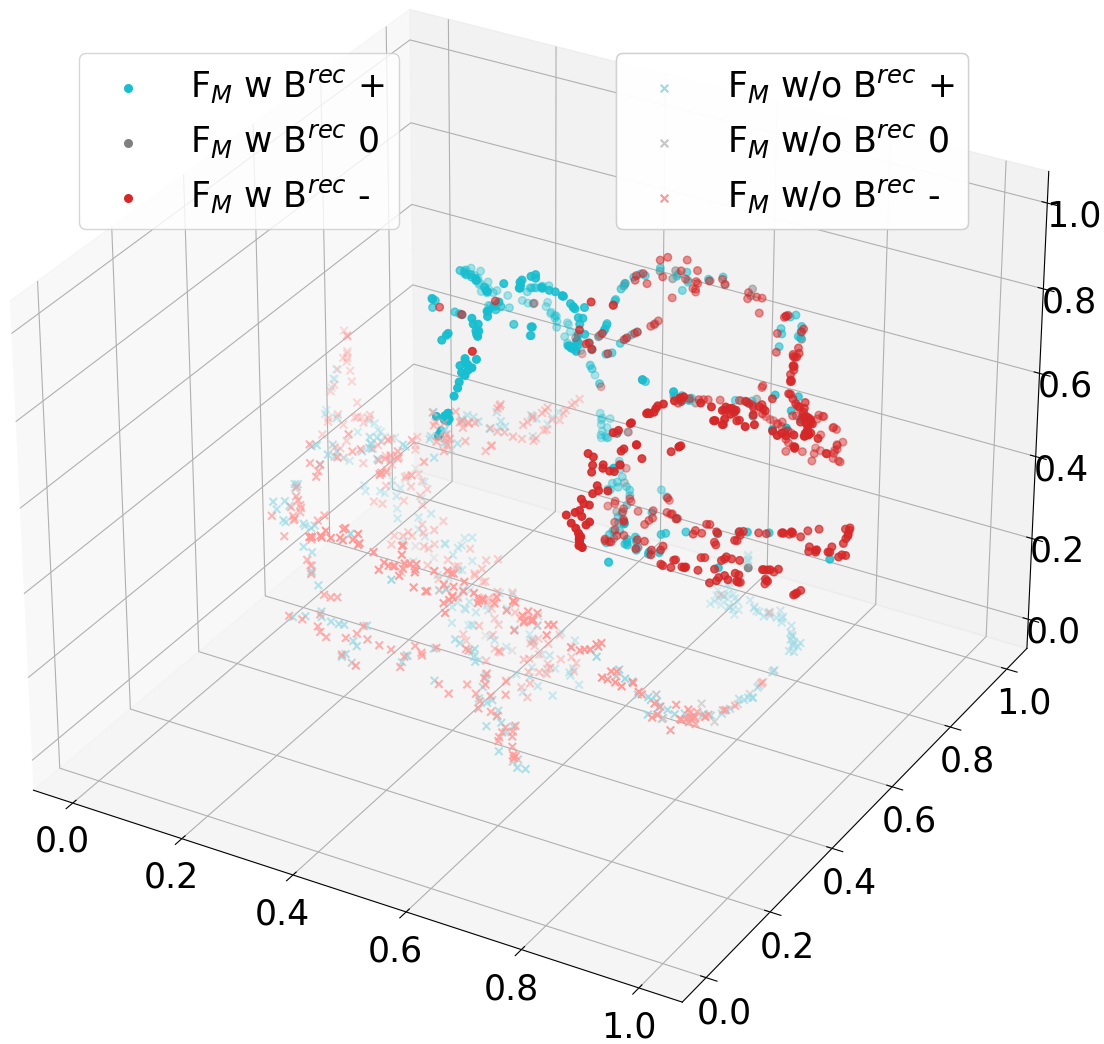} 
    % {figs/multimodal_latent_MOSI.pdf} 
 \label{vis_latent}
 \vspace{0.1cm}
 \end{minipage}
 }
 \hspace{-0.4cm}
 \subfigure[Case Study on MOSI (\#1-2) and IEMOCAP (\#3-4)]{
 \begin{minipage}[bthp]{0.62\linewidth}
    \setlength\tabcolsep{8pt}
    \hspace{-0.3cm}
    \scalebox{0.53}{
  \begin{tabular}{ccccccc}
    \toprule[1.5pt]
    \# & Multimodal Input ($u\in\{language,vision,audio\}$) & \makecell[c]{Modal\\Status} & \multicolumn{1}{c}{\makecell[c]{Ground\\Truth}} & \makecell[c]{Predict\\w CyIN} & \makecell[c]{Predict\\w/o CyIN} \\
    \midrule[1.5pt]
    \multirow{3}{*}[-13pt]{1} & \makecell[c]{``And he, I don't know, he maybe got mad when hah I don't know"} & \XSolidBrush & \multirow{3}{*}[-17pt]{\colorbox{red!30}{-0.250}} & \multirow{3}{*}[-17pt]{\colorbox{red!30}{-0.364}} & \multirow{3}{*}[-17pt]{\colorbox{green!20}{0.263}} \\
    & \makecell[c]{\includegraphics[width=2.0cm]{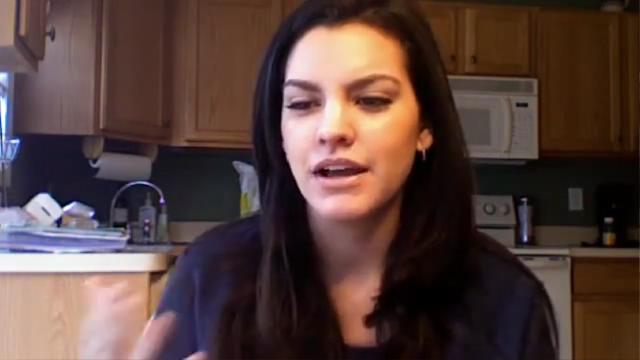} \includegraphics[width=2.0cm]{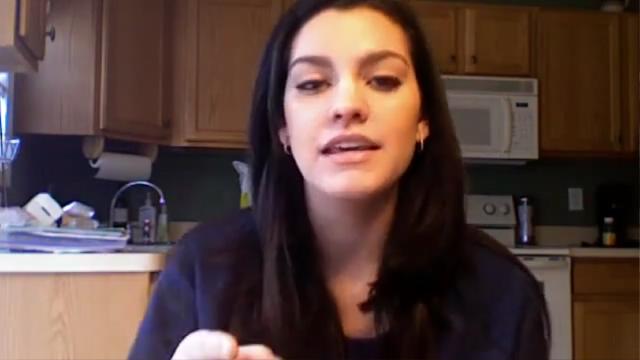}
    \includegraphics[width=2.0cm]{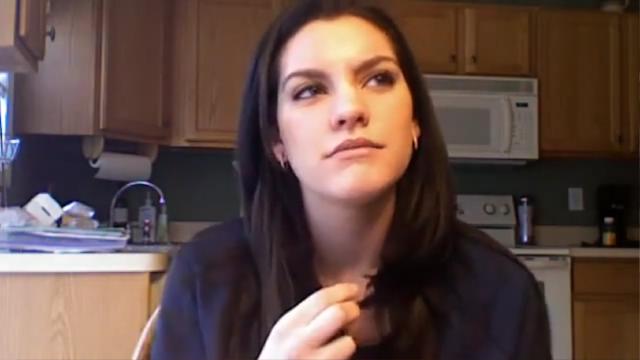}
    \includegraphics[width=2.0cm]{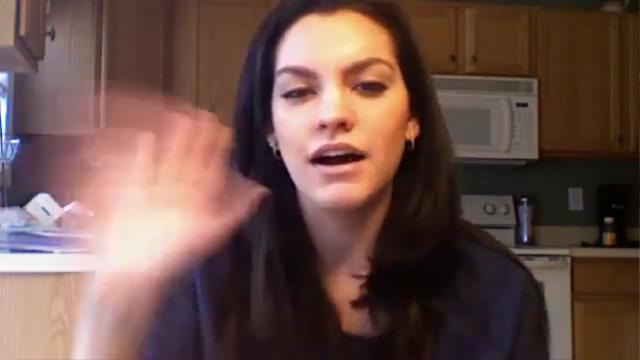}} & \Checkmark &  & & \\
    & Speculative and Wondering Tone & \Checkmark &  & & \\
    \midrule[0.5pt]
    
    \multirow{3}{*}[-15pt]{2} & ``So I think it was cool to actually see Ray Park in action, great action star" & \Checkmark & \multirow{3}{*}[-17pt]{\colorbox{green!50}{2.400}} & \multirow{3}{*}[-17pt]{\colorbox{green!50}{2.609}} & \multirow{3}{*}[-17pt]{\colorbox{green!30}{1.110}}  \\
    & \makecell[c]{\includegraphics[width=2.0cm]{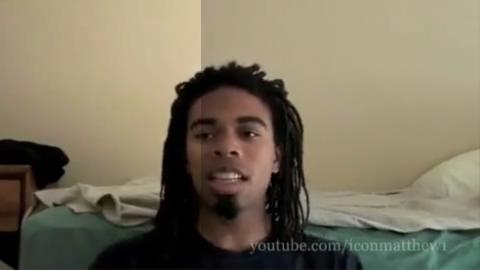} \includegraphics[width=2.0cm]{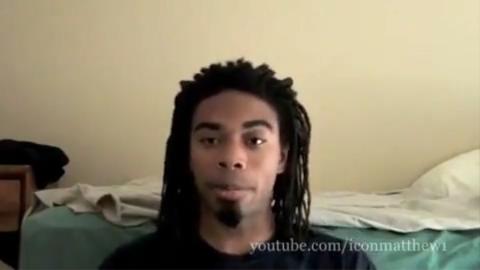}
    \includegraphics[width=2.0cm]{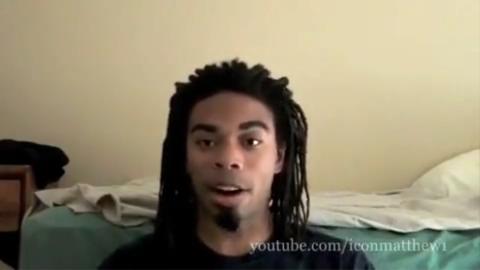}
    \includegraphics[width=2.0cm]{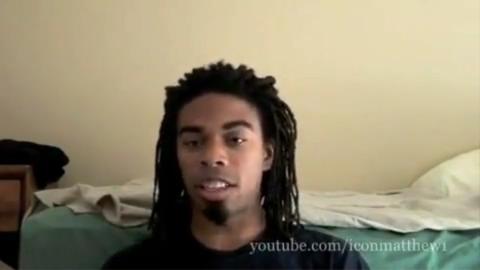}} & \XSolidBrush & & & \\
    & Satisfied and Relief Tone & \XSolidBrush & & & \\
    \midrule[0.5pt]
    
    \multirow{3}{*}[-17pt]{3} & \makecell[c]{``You needn't be so grand simply because you don't happen to \\ want any at the moment.."} & \XSolidBrush &\multirow{3}{*}[-17pt]{\colorbox{red!20}{Angry}} & \multirow{3}{*}[-17pt]{\colorbox{red!20}{Angry}} & \multirow{3}{*}[-17pt]{\colorbox{cyan!20}{Frustrated}} \\
    & \makecell[c]{\includegraphics[width=2.0cm]{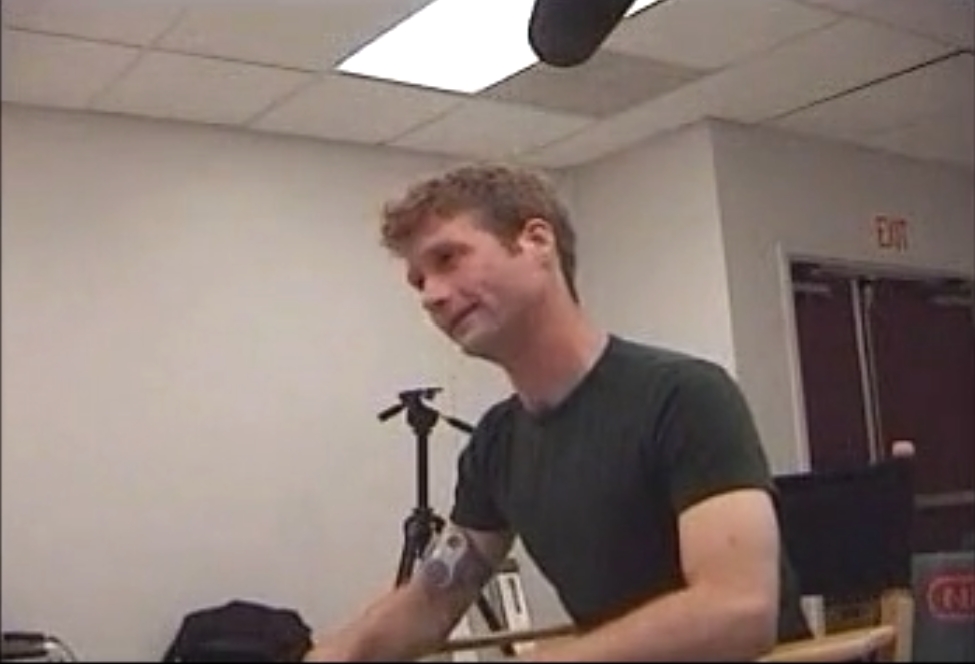} \includegraphics[width=2.0cm]{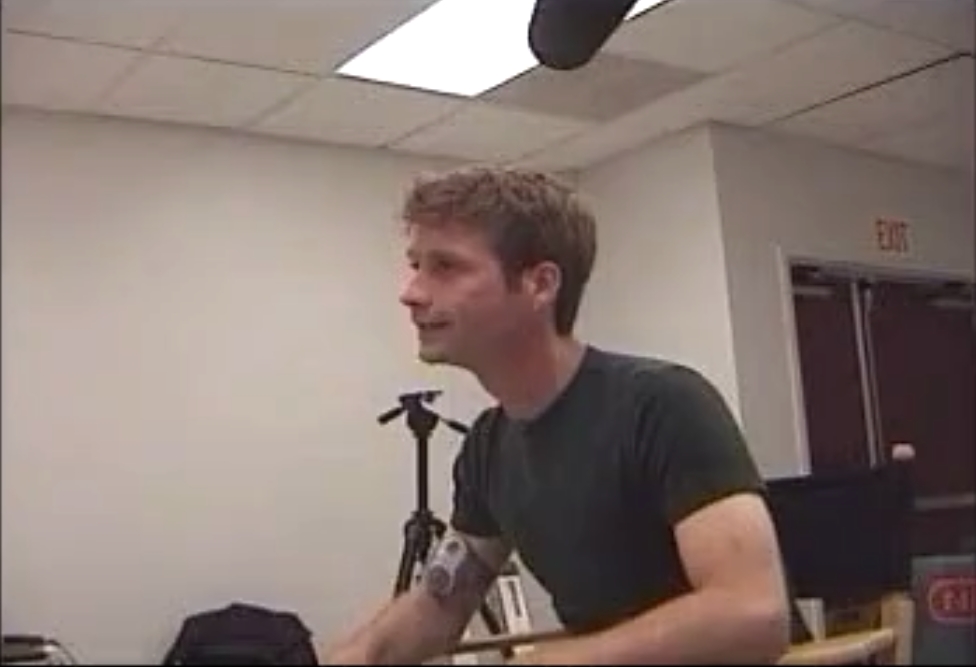}
    \includegraphics[width=2.0cm]{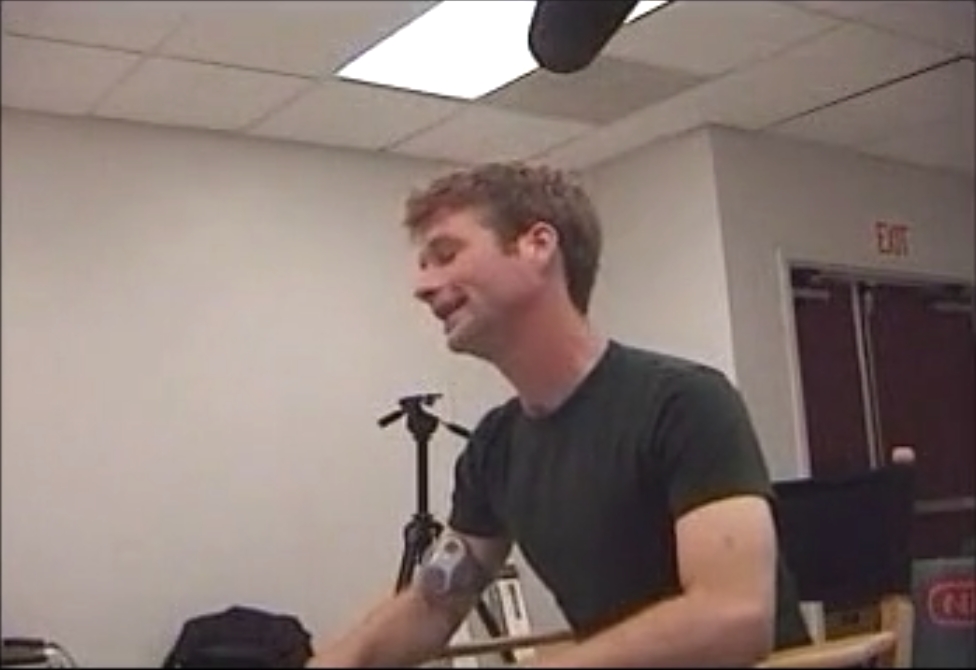}
    \includegraphics[width=2.0cm]{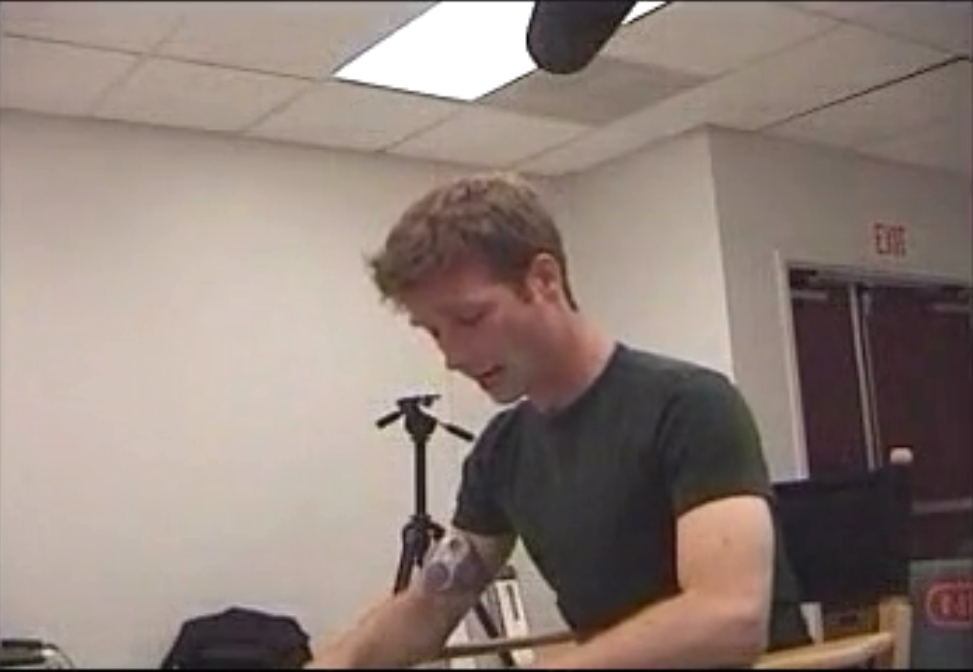}} & \Checkmark & & & \\
    & Accusatory and Agitated Tone & \Checkmark & & & \\
    \midrule[0.5pt]

    \multirow{3}{*}[-17pt]{4} & \makecell[c]{``You won't have trouble, You won't have trouble."} & \Checkmark &\multirow{3}{*}[-17pt]{\colorbox{orange!20}{Happy}} & \multirow{3}{*}[-17pt]{\colorbox{orange!20}{Happy}} & \multirow{3}{*}[-17pt]{\colorbox{gray!20}{Neutral}} \\
    & \makecell[c]{\includegraphics[width=2.0cm]{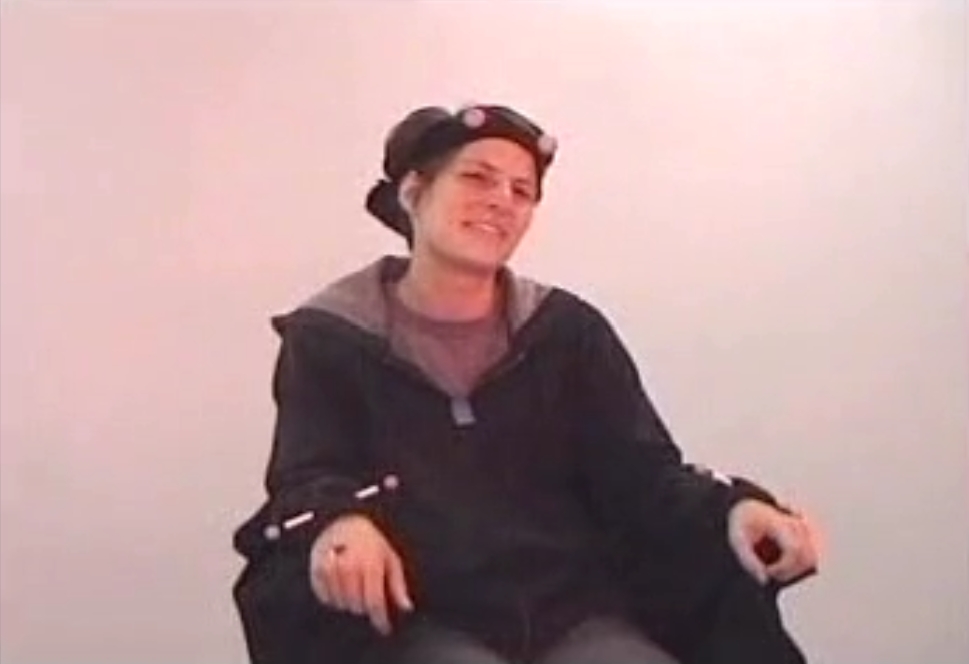} \includegraphics[width=2.0cm]{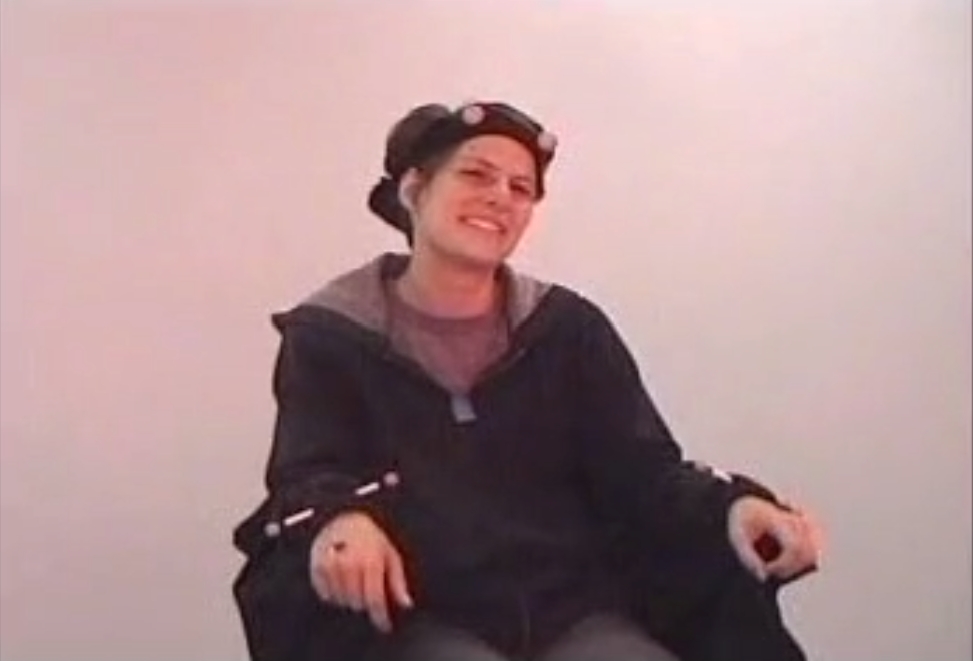}
    \includegraphics[width=2.0cm]{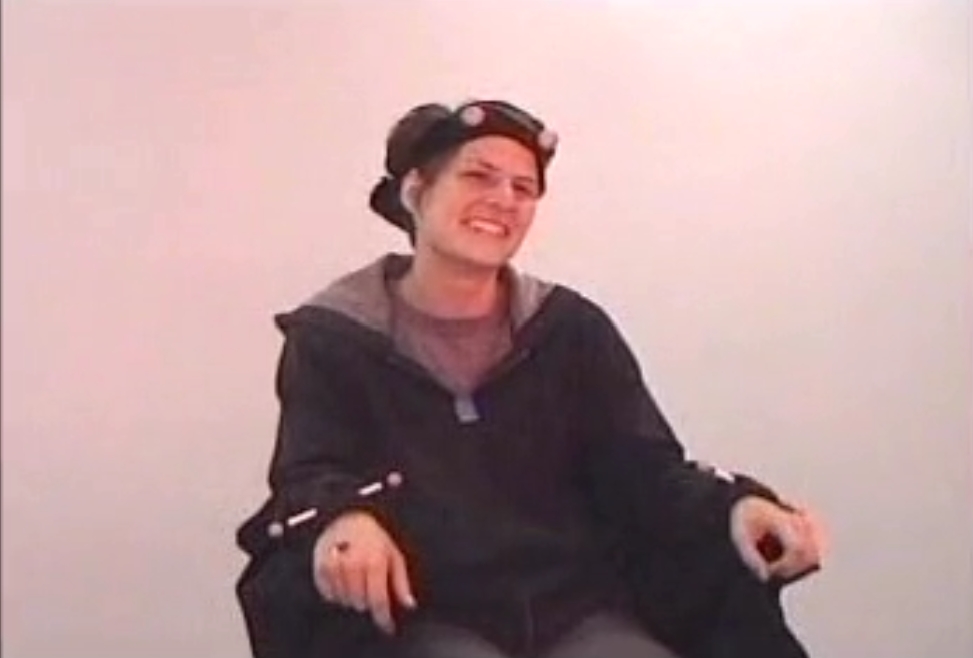}
    \includegraphics[width=2.0cm]{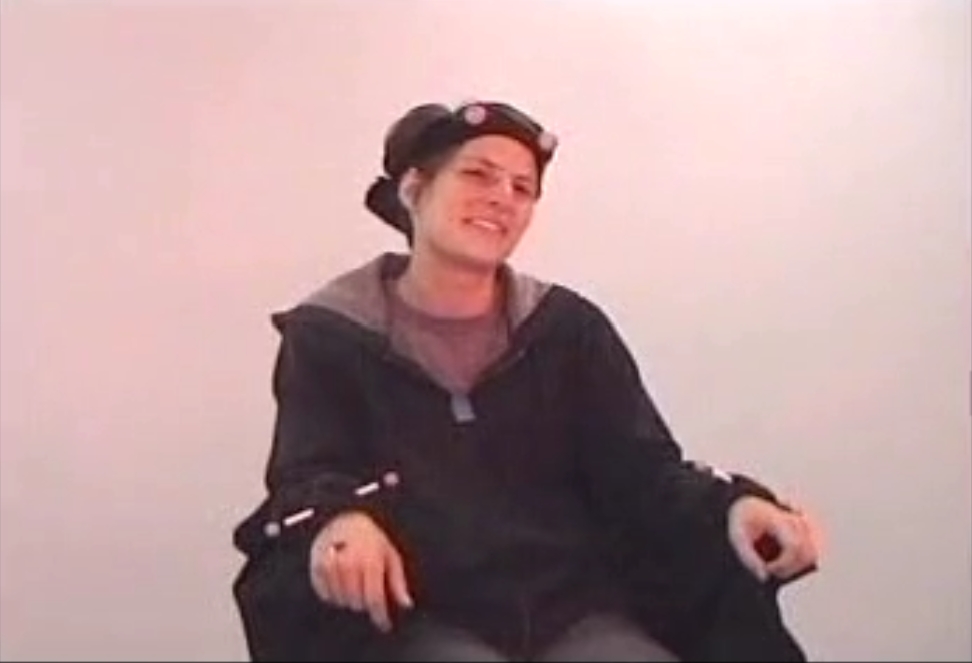}} & \XSolidBrush & & & \\
    & Calm and Fast Tone & \Checkmark & & & \\
    \bottomrule[1.5pt]
  \vspace{0.1cm}

  \end{tabular}
  }
 \label{table_case}
 \end{minipage}
 }
\vspace{-0.1cm}
\caption{(a) Feature distribution of translated unimodal latents and multimodal latents and (b) Examples on the test set of MOSI and IEMOCAP datasets when inferring with and without reconstructed information from CyIN. The \Checkmark and \XSolidBrush in modal status denotes the remained and missing modalities.}
\end{figure*}

\textbf{Qualitative Comparisons.}
We project the features from test set of MOSI dataset into a 3D t-SNE space \cite{hinton2008visualizing}. As shown in Figure \ref{vis_latent}, we firstly visualize the distribution of translated $B^{rec}_u$ and original unimodal latents $B_u$ to show the reconstruction quality. The cross-modal translated latents closely clustered to the original one, illustrating the effectiveness of CyIN in transferring information across various modalities despite the huge modality gap.

Besides, we present multimodal latents $F_M$ with and without reconstructed bottleneck $B^{rec}$ with random missing $MR=0.7$. We can observe that inferring with missing modalities leads to serious interference to the multimodal representations, while the reconstructed latents can competently supplement the missing information during multimodal fusion, yielding better clustering result. 

\textbf{Case Study.}
As shown in Figure \ref{table_case}, we compare model predictions with and without CyIN on some examples from test set of MOSI and IEMOCAP datasets under fixed missing modality protocol. In samples$ \#1$ and $\#3$, removing the semantic utterance causes the pretrained model to make incorrect, even opposite, predictions, which implicitly showing the dominant role of language modality in multimodal affective computing tasks \cite{pham2019found,hazarika2020misa,lin2023dynamically,zhang2024towards}. On the other hand, the presence of other modalities also have delicate contribution in the accuracy of prediction, referring to the subjective biased prediction when missing audio or vision modality in sample $\#2$ and $\#4$. In contrast, by reconstructing the missing information with the proposed CyIN, predictions become more stable and precise, demonstrating the superiority and robustness of the informative bottleneck space.

\textbf{Ablation Study.} We conduct ablation study with the proposed modules on MOSI and IEMOCAP datasets with complete and the most severe random missing protocol $MR=0.7$, as shown in Table \ref{table_ablation}. Token- and label-level IB are both crucial for constructing the informative latent space, as removing either results in performance degradation. The cross-modal cyclic interaction and translation mainly effect in capturing modality-shared features in multimodal fusion and enhancing incomplete reconstruction, respectively. Moreover, we demonstrate the efficiency of informative bottleneck space in both complete and incomplete multimodal learning. Without the constrain of bottleneck, the task-irrelevant redundancy and heterogeneous noise in the original feature space raises difficulty in integrating information from various modalities and reconstructing missing modalities. Lastly, reconstructed from the remained modalities, the translated latents can productively increase the robustness to missing modality issue especially in severe incomplete input circumstance.

\begin{table}[htbp]
\centering
\setlength\tabcolsep{15pt}
\caption{Ablation study of the proposed CyIN on MOSI and IEMOCAP dataset with complete multimodal settings $u\in\{l,a,v\}$ and random missing protocols with missing rates $MR=0.7$.}
\label{table_ablation}
\scalebox{0.73}{
\begin{tabular}{cccccccc}
\toprule[1.5pt]
\multirow{2}{*}{Setting} & \multirow{2}{*}{Model Variants} & \multicolumn{4}{c}{MOSI} & \multicolumn{2}{c}{IEMOCAP} \\
\cmidrule(r){3-6}\cmidrule(r){7-8}
& & Acc7$\uparrow$ & F1$\uparrow$ & MAE$\downarrow$ & Corr$\uparrow$ & Acc$\uparrow$ & wF1$\uparrow$\\

\midrule[1.5pt]
    \multirow{6}{*}{\makecell[c]{Complete\\{$u\in\{l,a,v\}$}}} ~ & \textbf{CyIN}  & \textbf{48.0} & \textbf{86.3} & 0.712 & \textbf{0.801} & \textbf{66.1} & \textbf{66.0} \\ 
    & w/o $\mathcal{L}_{tib}$ & 43.9 & 84.3 & 0.737 & 0.789 & 65.2 & 64.8 \\
    & w/o $\mathcal{L}_{lib}$ & 47.3 & 85.4 & \textbf{0.693} & 0.800 & 63.6 & 63.9 \\
    & w/o Cyclic Interaction & 43.4 & 85.9 & 0.742 & 0.795 & 64.3 & 63.6 \\
    & w/o Cyclic Translation & 43.3 & 85.7 & 0.743 & 0.797 & 65.2 & 65.0 \\
    & w/o Informative Space & 42.0 & 83.1 & 0.747 & 0.782 & 62.3 & 62.1 \\
\midrule[0.5pt]
    \multirow{7}{*}{\makecell[c]{Random\\Missing\\$MR=0.7$}} ~ &\textbf{CyIN} & \textbf{28.0} & 65.9 & \textbf{1.117} & \textbf{0.530} & \textbf{48.6} & \textbf{49.0} \\ 
    & w/o $\mathcal{L}_{tib}$ & 27.3 & 63.5 & 1.223 & 0.509 & 46.3 & 46.2 \\
    & w/o $\mathcal{L}_{lib}$ & 25.1 & 63.7 & 1.220 & 0.516 & 43.2 & 42.9 \\
    & w/o Cyclic Interaction & 26.5 & \textbf{67.8} & 1.134 & 0.473 & 47.8 & 48.6 \\
    & w/o Cyclic Translation & 24.5 & 63.9 & 1.171 & 0.473 & 47.1 & 46.6 \\
    & w/o Informative Space & 23.7 & 62.9 & 1.240 & 0.430 & 44.1 & 43.6 \\
    & w/o Translated Latents & 23.4 & 56.7 & 1.299 & 0.441 & 41.2 & 39.2 \\

\bottomrule[1.5pt]
\end{tabular}
}
\end{table}

\textbf{Hyper-parameter Sensitivity.}
We empirically evaluate the effect of various hyper-parameter settings of $\beta$ and $\gamma$ in loss function of Equ. \ref{equ_loss_total} under $MR=0.7$. The results in Table \ref{table_beta_gamma} shows that setting $\beta=8-32$ gives better performance as a balanced trade-off for mututal information between $I(S;B)$ and $I(B;T)$, where improper bottleneck strength harms performance. Besides, setting $\gamma=10$ yields the best overall performance across metrics, indicating that moderate reconstruction enforces better consistency across modalities. We observe that a small  $\gamma$ results in weak regularization, making it hard to align modalities under severe missing conditions. In contrast, a large $\gamma$ hurts performance by focusing too much on consistency and ignoring the need to effectively build informative space.

Moreover, to ensure generality and stability under diverse missing modality scenarios, CyIN assumes each modality have equal contribution in multimodal learning. However, in practice, different modalities may contribute unequally depending on the task \cite{peng2022Balanced}. For instance, language plays a dominant role in multimodal sentiment analysis, as highlighted by prior works \cite{lin2023mtmd,lian2024merbench,zhang2024towards}. To partially address this, we conduct experiments by varying the number of CRA layers in the cross-modal translation modules in Table \ref{table_cra_layer}. Specifically, we allocated more layers to reconstruct the language modality comparing with audio and vision modalities, reflecting its higher importance and allowing the model to learn representation with more semantics. This design show the flexibility of CyIN architecture without hardcoding any modality-specific priors in it. 

\begin{table}
\begin{minipage}[htbp]{0.45\textwidth}
\centering
\caption{Hyper-parameter sensitivity on $\beta$ and $\gamma$ of CyIN on random missing protocols with missing rate $MR=0.7$.}
\label{table_beta_gamma}
\scalebox{0.8}{
\begin{tabular}{cccccc}
\toprule[1.5pt]
\multicolumn{2}{c}{\multirow{2}{*}{\makecell{Model Variants}}} & \multicolumn{4}{c}{MOSI} \\
\cmidrule(r){3-6}
& & Acc7$\uparrow$ & F1$\uparrow$ & MAE$\downarrow$ & Corr$\uparrow$ \\ 
\midrule[0.5pt]
   \multirow{6}{*}{$\gamma$=10} & $\beta$=2  & 24.6 & 59.4 & 1.248  & 0.461  \\ 
    & $\beta$=4  & 27.5 & 61.3 & 1.226 & 0.478 \\ 
    & $\beta$=8  & \textbf{30.0} & 63.6 & 1.199 & 0.519 \\ 
    & $\beta$=16  & 28.0 & 65.9 & \textbf{1.117} & \textbf{0.530} \\  
    & $\beta$=32  & 26.4 & \textbf{66.0} & 1.118 & 0.506 \\ 
    & $\beta$=64  & 23.3 & 64.1 & 1.206 & 0.463 \\
\midrule[0.5pt]
    \multirow{5}{*}{$\beta$=16} & $\gamma$=1  & 25.8 & 64.1 & 1.197 & 0.437 \\ 
    & $\gamma$=5  & 27.5 & 63.1 & 1.135 & 0.488 \\ 
    & $\gamma$=10  & \textbf{28.0} & \textbf{65.9} & \textbf{1.117} & \textbf{0.530} \\  
    & $\gamma$=15  & 26.8 & 62.9 & 1.272 & 0.519 \\ 
    & $\gamma$=20  & 25.5 & 60.4 & 1.311 & 0.490 \\
\bottomrule[1.5pt]
\end{tabular}
}
\end{minipage}
\hfill
\begin{minipage}[htbp]{0.45\textwidth}
\centering
\caption{Hyper-parameter sensitivity on different ratios of CRA layer for translation among language, vision, audio modalities of CyIN on MOSI dataset on random missing protocols with missing rate $MR=0.7$.}
\label{table_cra_layer}
\scalebox{0.9}{
\begin{tabular}{ccccc}
\toprule[1.5pt]
\multirow{2}{*}{\makecell[c]{$\#$ Layer\\($la$:$lv$:$av$)}} & \multicolumn{4}{c}{MOSI} \\
\cmidrule(r){2-5}
& Acc7$\uparrow$ & F1$\uparrow$ & MAE$\downarrow$ & Corr$\uparrow$ \\

\midrule[0.5pt]
    1:1:1  & 28.0 & 65.9 & 1.117 & 0.530  \\
    2:2:1  & 28.5 & 67.2 & 1.122 & 0.521 \\ 
    4:4:1  & \textbf{28.7} & \textbf{67.5} & \textbf{1.114} & 0.532 \\ 
    1:1:2  & 27.5 & 66.3 & 1.119 & 0.531 \\ 
    1:1:4  & 27.4 & 65.6 & 1.116 & \textbf{0.534} \\ 
\bottomrule[1.5pt]
\end{tabular}
}
\end{minipage}
\end{table}

\textbf{Computational Efficiency.} 
Compared with the state-of-the-art methods, the proposed CyIN achieves the lowest total parameter count and significantly lower FLOPs as shown in Table \ref{table_computation_efficiency}. The inference time of CyIN is $\sim3\times$ faster than GCNet and $\sim5\times$ faster than IMDer. Due to the cyclic translation process during training, the training time of CyIN is slightly slower than GCNet which constructs graph neural networks for modality reconstruction, but still faster then IMDer which utilizes score-based diffusion model for cross-modal generation.

\begin{table*}[thbp]
\centering
\setlength\tabcolsep{7pt}
\caption{Comparion of training and inference computation efficiency on MOSI dataset. }
\vspace{0.1cm}
\label{table_computation_efficiency}
\scalebox{0.9}{
% \begin{tabular}{*{14}{c}}
\begin{tabular}{ccccc}
\toprule[1.5pt]
Model &  Total Param (M) & Total Training Time & Inference Time (/iteration)  & FLOPs (T) \\
\midrule[0.5pt]
GCNet & 144.34 M & \textbf{1.47h} & 70.62s & 3.747 \\
IMDer & 168.31 M & 1.89h  & 103.75s & 5.466 \\
\midrule[0.5pt]
\textbf{CyIN (ours)} & \textbf{123.49 M} & 1.61h & \textbf{22.41s} & \textbf{1.594} \\
\bottomrule[1.5pt]
\end{tabular}
}
\vspace{-1em}
\end{table*}

\section{Conclusion}
In this paper, we present a novel framework named CyIN, which constructs an informative latent space to jointly conduct complete and incomplete multimodal learning. Guided by token- and label-level Information Bottlenecks, CyIN succeeds in learning a compact yet semantically rich bottleneck latent which purifies task-related features and improves robust multimodal fusion. The proposed cyclic interaction and translation mechanism further encourage sufficient exploration in inter-modal dynamics and enhances the reconstruction quality in missing modalities. Comprehensive experiments on 4 multimodal datasets demonstrate that CyIN not only surpass previous methods in complete multimodal learning but also retains superior performance and stability under various incomplete scenarios, highlighting its effectiveness and generalization capacity for real-world multimodal applications.

\section*{Limitations}
Our framework still has several limitations. 1) It treats all modalities equally to improve the generalization, while ignoring possible imbalance contribution from various modalities. We will try to adaptively allocate weights according to the information abundance of each modality. 2) The performance of current cross-modal translators can be replaced by more recent generative approaches such as diffusion models or flow-based models to attain optimal reconstruction performance. 3) We suppose broader training on diverse multimodal understanding datasets or integration with Large Language Models could strengthen the generalization and scalability of the proposed framework. 

% lacks mechanisms to modulate latent representations based on semantic attributes like sentiment or relevance, and
% ordinal modulation of informative latent space according to specific labels with sentiment, relevance, ...

% imbalance contribution from various modalities to allocate various weight with information abundance

% utilize new generative or reconstruction based network to replace the translator such as diffusion or rectified flow

% try more datasets and combine with MLLM

\section*{Acknowledgments}
% \begin{ack}
This work was supported by the National Natural Science Foundation of China (62076262, 61673402, 61273270, 60802069) and by the International Program Fund for Young Talent Scientific Research People, Sun Yat-Sen University.

% Do {\bf not} include this section in the anonymized submission, only in the final paper. You can use the \texttt{ack} environment provided in the style file to automatically hide this section in the anonymized submission.
% \end{ack}

% \section*{References}

% References follow the acknowledgments in the camera-ready paper. Use unnumbered first-level heading for
% the references. Any choice of citation style is acceptable as long as you are
% consistent. It is permissible to reduce the font size to \verb+small+ (9 point)
% when listing the references.
% Note that the Reference section does not count towards the page limit.

% \newpage


{
\bibliographystyle{unsrtnat}
\bibliography{custom}
}

%%%%%%%%%%%%%%%%%%%%%%%%%%%%%%%%%%%%%%%%%%%%%%%%%%%%%%%%%%%%

%%%%%%%%%%%%%%%%%%%%%%%%%%%%%%%%%%%%%%%%%%%%%%%%%%%%%%%%%%%%

\newpage
\appendix

\section{Related Works}
\label{appendix_relatedwork}
\subsection{Multimodal Representation Learning}
Multimodal representation learning aims at constructing multimodal system to conduct multimodal understanding tasks including multimodal sentiment analysis \cite{hazarika2020misa,han2021improving,lin2023mtmd,zhang2023learning,xiao2024neuroinspired} or emotion recognition \cite{hu2022unimse,li2023joyful,lin2024semanticmac,cheng2024emotionllama}, multimodal recommendation \cite{li2025generating} and multimodal segmentation \cite{pipoli2025imfuse} and so on. The common framework of multimoda learning can be divided into two main modules: modality-specific representation learning and multimodal fusion, according to the purpose of reflecting heterogeneity and interconnections between modality elements \cite{liang2024foundations}. 
% video understanding \cite{he2024malmm,zeng2025timesuite}

On the one hand, effective extraction of task-relevant features within each individual modality forms the foundation of multimodal representation learning \cite{lian2024merbench,lin2024semanticmac}. Across various modalities, these modality-specific encoders emphasize critical features and suppress unrelated noise, thereby shortening the path from raw data to semantically meaningful unimodal representations. On the other hand, multimodal fusion aims to bridge modality gaps and exploit inter-modal complementarity \cite{gkoumas2021makes,liang2024foundations}. Early fusion schemes simply concatenate unimodal representations  while late-fusion approaches combine unimodal predictions via weighted voting or gating. Hybrid fusion combine early and late fusion to conduct hierarchical and more delicated cross-modal interaction \cite{zhao2024deep}.

With the emerge of attention mechanism \cite{vaswani2017attention}, multimodal model with Transformer-based architecture efficiently capture intra- and inter-modal dynamics by interleaving unimodal representation with self‐ and cross‐modal attention layers \cite{wang2023large,xu2023multimodal}. However, most of previous multimodal methods assume paired, fully observed modalities at both training and inference stages to achieve the state-of-the-art performance of multimodal understanding.

\subsection{Missing Modality Issue}
The missing modality issue occurs when pre-trained multimodal models suffer from severe performance degradation at downstream inference due to the absent input of modality data \cite{lin2023missmodal,huan2024unimf}. In real world, the presence of multimodal input could not be guaranteed due to numerous reasons, such as sensor failure, hardware malfunctions, privacy limitations, environmental disruptions, and data transmission problems
\cite{wu2024deep}. Empirical analysis have shown that the efficacy of multimodal methods critically depends on complete presence of modalities \cite{hazarika2022analyzing}, and that Transformer-based encoders are especially sensitive to missing inputs \cite{ma2022multimodal}. Thus, addressing missing modality issue is essential for robust performance in real-world applications, which is named as incomplete multimodal learning.
%Currently most state-of-the-art multimodal large-language models are built upon Transformer-based architectures \cite{xu2023multimodal,wang2023large}. 

Taking alternative perspective of incomplete multimodal learning, the missing modality issue can be regarded as missing specific-view information issue in the multi-view learning \cite{tang2024incomplete}. Previous methods mainly focus on aligning complete and incomplete multimodal representations \cite{zhang2022cpmnet,poklukar2022geometric} or generating the missing information with incomplete multimodal input \cite{zhao2021missing,lian2023gcnet,wang2023incomplete}. While alignment‐based models avoid the complexity of explicit generation, they often yield suboptimal inference with uncertain and heterogeneous information asymmetry. Diversely, generative-based models remain susceptible to the task-unrelated noise and massive redundancy in the unimodal features, which can easily destabilize the information reconstruction or generation process. 
%CPM-Nets\cite{zhang2022cpmnet}  balance the consistency and complementarity of representation across different missing-views in a common latent space and GMC \cite{poklukar2022geometric} employs geometric multimodal contrastive learning to gather unimodal and multimodal representations.
% MMIN \cite{zhao2021missing} introduces forward and backward imagination networks to transfer information between available and absent modalities , and GCNet \cite{lian2023gcnet} employs graph-completion networks to infer missing temporal or speaker features prior in multimodal fusion. More recent generative models such as diffusion models are used to synthesize missing modalities with an auxiliary denoising network such as IMDer does \cite{wang2023incomplete}

Moreover, previous methods require adjusting according to specific missing scenarios to achieve optimal performance, restricting their applications in the uncertain circumstance of the real world. Beside, most of them sacrifice performance of complete multimodal learning to increase the robustness to the missing modality issue. Therefore, achieving a single unified framework that jointly optimize both complete and incomplete multimodal learning remains an open challenge.

\subsection{Deep Learning with Information Bottleneck}
Original introduced by \citet{tishby1999information}, Information Bottleneck (IB) formulates representation learning as an information‐theoretic approach that defines  task-relevant representation as trade‐off between feature compression and task prediction \cite{michael2018on}. Due to the excellent interpretability of representation learning and generalization estimation of the deep neural networks \cite{tishby2015deepib,shwartz2017opening,icml2023howdoesib}, IB has gained continuous interests in deep learning, especially in multi-view problem \cite{wang2019deep,federici2020Learning,Wan2021multiview}. VIB \cite{alemi2017vib} introduce variational approximation to generate the bottleneck latents in the information flow, sharing similar objective form with VAE \cite{kingma2014vae} in generation. DeepIMV \cite{lee2021deepimv} utilize product-of-experts to integrate the marginal specific-view representation into a joint latent representation. FactorCL \cite{liang2023factorized} factorizes task-relevant information into shared and unique information with multi-view redundancy defined by mutual information. Nevertheless, they can neither be employed to varying missing circumstance nor succeed in achieving optimal intra- and inter-modal information extraction performance. 

In this paper, we adopt the principles of IB to construct an effective informative latent space by designing information flow to features among modalities and guidance of semantics, paving the way for enhanced performance in both complete and incomplete multimodal learning.

\section{Derivations of the Variational Information Bottleneck}
\label{appendix_vib}
In this section, we derive the formula of the original variational information bottleneck and the proposed token- and label-level information bottleneck in detail, as shown in Equ. \ref{equ_vib}, Equ. \ref{equ_token_ib2} and Equ. \ref{equ_label_ib_final1}-\ref{equ_label_ib_final2}. Besides, we provide the differentiable sampling process of the informative latents presented in Equ. \ref{equ_reparameterization}.

Equ. \ref{equ_vib} represents the information bottleneck loss $\mathcal{L}_{ib}$ for the information flow as ($S\rightarrow B\rightarrow T$), where $S$ denotes the source state, $B$ denotes the bottleneck latents and $T$ denotes the target state. Recall that the loss is firstly given by the constrained of mutual information, formulated as:
\begin{equation}
    \min \mathcal{L}_{ib}(S,T) = \min_{p(B|S)} I(S;B) - \beta\ I(B;T)
\end{equation}
where $\beta>0$ is a trade-off parameters to balance the two mutual information terms. Directly obtaining the mutual information among $S/B/T$ is unachievable. Therefore, we tend to obtain the upper bound of $\mathcal{L}_{ib}(S,T)$ to transfer the objective minimization problem to an evidence upper bound optimization problem.

Starting with the first term $I(S;B)$, writing is out in full form as:
\begin{equation}
\begin{aligned}
    I(S;B) &= \int dB\ dS\ p(S, B) \log{\frac{p(S, B)}{p(S)p(B)}} = \int dB\ dS\ p(S, B) \log{\frac{p(B|S)}{p(B)}} \\
    & = \int dB\ dS\ p(S, B) \log{p(B|S)} - \int dB\ dS\ p(B) \log{p(B)} 
\end{aligned}
\end{equation}

Considering that the computation of marginal distribution of the bottleneck latents $p(B)=\int{dS p(B|S)p(S)}$ might be difficult, let $q(B)$ be a variational approximation to this marginal distribution. With the non-negative Kullback Leibler (KL) divergence $KL(p(B)\parallel q(B))\geq 0$, we have $\int dB\ p(B) \log{p(B)} \geq \int dB\ p(B) \log{q(B)}$, the following upper bound can be derived:
\begin{equation}
\begin{aligned}
    I(S;B) &\leq \int dBdS\ p(S, B) \log{p(B|S)} - \int dB\ \ p(B) \log{q(B)} \\
    &=\int dBdS\ p(S)p(B|S) \log{\frac{p_\theta(B|S)}{q(B)}} 
\end{aligned}
\end{equation}
where $p_\theta(B|S)$ can be learned by IB encoder $E_S:S\mapsto B$.

Then for the second term $I(B;T)$, consider it in the same full form as:
\begin{equation}
    I(B;T)= \int dT\ dB\ p(T,B) \log{\frac{p(T, B)}{p(T)p(B)}} = \int dT\ dB\ p(T, B) \log{\frac{p(T|B)}{p(T)}}
\end{equation}

Since $p(T|B)$ is intractable, introducing IB decoder $D_T:B\mapsto T$, let $q_\phi(T|B)$ be a variational approximation to $p(T|B)$. Similarly, with KL divergence $KL(p(T|B)\parallel q_\phi(T|B))\geq 0$, we have $\int dT\ p(T|B) \log{p(T|B)} \geq \int dT\ p(T|B) \log{q_\phi(T|B)}$. Hence $I(B;T)$ has a lower bound as follows:
\begin{equation}
\begin{aligned}
    I(B;T) &\geq \int dT\ dB\ p(T, B) \log{\frac{q_\phi(T|B)}{p(T)}} \\
    &=\int dT\ dB\ p(T, B) \log{q_\phi(T|B)}-\int dT\ p(T) \log{p(T)} \\
    &=\int dT\ dB\ p(T, B) \log{q_\phi(T|B)}+H(T)
\end{aligned}
\end{equation}
where $H(T)$ is the entropy of target state $T$, determined only by the distribution of $T$ itself, no matter as the unimodal token embeddings or ground truth labels. Thus, $H(T)$ can be dropped in the loss function. When the source state $S$ is diverse from the target state $T$, we introduce the relationship that $B$ is independent of $T$ given $S$:
\begin{equation}
    p(T,B)=\int dS\ p(S,T,B) = \int dS\ p(S) p(T|S) p(B|S) 
\end{equation}

Note that when $S=T$, we have $p(T,B)=p(S,B)=\int p(S)p(B|S)=\int p(S)p(T|S)p(B|S)$ where above expression still stand. Besides, $p(T|S)$ denotes the known distribution for the joint and paired data samples. Utilizing data samples within each batch as the empirical data distribution, we have $p(S,T)=p(S)p(T|S)=\frac{1}{N}\sum_{n=1}^N\delta_{S_n}(S)\delta_{T_n}(T)$. Hence, $I(B;T)$ can be generally denoted as:
\begin{equation}
\begin{aligned}
    I(B;T) &\geq \int dS\ dT\ dB\ p(S) p(T|S) p(B|S) \log{q_\phi(T|B)} \\
    &\approx \mathbb{E}_{S\sim p(S)}\left[\int dB\ p(B|S) \log{q_\phi(T|B)}\right]
\end{aligned}
\end{equation}

% After all, computing $I(B;T)$ only requires samples from the IB encoder $E_S:S\mapsto B$ and the tractable variational approximation $q_\phi(T|B)$.

Combining the bound of $I(S;B)$ and $I(B;T)$, we can derive the following upper bound for $\mathcal{L}_{ib}$ as Equ. \ref{equ_vib}:
\begin{equation}
\begin{aligned}
    & I(S;B) - \beta\ I(B;T) \\
    &\leq \int dBdS\ p(S)p(B|S) \log{\frac{p_\theta(B|S)}{q(B)}} - \beta\ \mathbb{E}_{S\sim p(S)}\left[\int dB\ p(B|S) \log{q_\phi(T|B)} \right]\\
    &= \mathbb{E}_{S\sim p(S)} \left[KL( p_\theta(B|S)\parallel q(B))\right] - \beta\ \mathbb{E}_{B\sim p(B|S)}\mathbb{E}_{S\sim p(S)} [\log q_\phi(T|B)]
\end{aligned}
\end{equation}

Finally, the optimization objective of variational information bottleneck can be rewritten as:
\begin{equation}
    \min\mathcal{L}_{ib} = \min\mathbb{E}_{S\sim p(S)} \left[KL( p_\theta(B|S)\parallel q(B))\right] - \beta\max\ \mathbb{E}_{B\sim p(B|S)}\mathbb{E}_{S\sim p(S)} [\log q_\phi(T|B)]
\end{equation}

Now, we derive the practical expression of two minimization and maximization terms of the above upper bound. 

First for minimizing $KL( p_\theta(B|S)\parallel q(B))$, in practice, we utilize standard Gaussian Distribution $\mathcal{N}(0,1)$ as the prior distribution of $q(B)$ and model the approximate posterior distribution $p_\theta(B|S)$ as a multivariate Gaussian distribution $\mathcal{N}(\mu, \varepsilon)$. Here $\mu$ and $\varepsilon$ denotes the mean and variance vectors of the latent Gaussian Distribution. Since each dimension of $q(B)$ and $p_\theta(B|S)$ are independent, we can derive the situation with one-dimensional Gaussian Distribution of $KL( p_\theta(B|S)\parallel q(B))$ as:
\begin{equation}
\begin{aligned}
    &KL( p_\theta(B|S)\parallel q(B)) = KL( \mathcal{N}(\mu_B, \sigma^2_B)\parallel\mathcal{N}(0,\textbf{I})) \\
    &= \int \mathcal{N}(\mu_B, \sigma^2_B) \log \frac{\mathcal{N}(\mu_B, \sigma^2_B)}{\mathcal{N}(0, \textbf{I})} \, db \\
    &= \int \mathcal{N}(\mu_B, \sigma^2_B) \log \frac{\frac{1}{\sqrt{2\pi \sigma^2_B}} \exp\left( -\frac{(b - \mu_B)^2}{2\sigma^2_B} \right)}{\frac{1}{\sqrt{2\pi}} \exp\left( -\frac{b^2}{2} \right)} \, db \\
    &= \int \mathcal{N}(\mu_B, \sigma^2_B) \left( -\log \sqrt{\sigma^2_B} - \frac{(b - \mu_B)^2}{2\sigma^2_B} + \frac{b^2}{2} \right) \, db \\
    &= -\frac{1}{2} \int \mathcal{N}(\mu_B, \sigma^2_B) \left(\log \sigma^2_B + \frac{(b - \mu_B)^2}{\sigma^2_B} - b^2 \right) \, db \\
    &= -\frac{1}{2} \left[ \log \sigma^2_B \int \mathcal{N}(\mu_B, \sigma^2_B) \, db + \frac{1}{\sigma^2_B} \int \mathcal{N}(\mu_B, \sigma^2_B)(b - \mu_B)^2 \, db - \int \mathcal{N}(\mu_B, \sigma^2_B) b^2 \, db \right]
\end{aligned}
\end{equation}

Since any probability density function integrates to 1 over its domain, we have $\int \mathcal{N}(\mu_B, \sigma^2_B) \, db=1$. According to the definition of the variance and second raw moment of Gaussian Distribution, we have $\int \mathcal{N}(\mu_B, \sigma^2_B)(b - \mu_B)^2 \, db=\sigma^2_B$ and $\int \mathcal{N}(\mu_B, \sigma^2_B) b^2 \, db=\mu^2_B+\sigma^2_B$. Thus, the analytical solution of $KL( \mathcal{N}(\mu_B, \sigma^2_B)\parallel\mathcal{N}(0,\textbf{I}))$ is computed as:
\begin{equation}
\begin{aligned}
    KL( \mathcal{N}(\mu_B, \sigma^2_B)\parallel\mathcal{N}(0,\textbf{I})) &=-\frac{1}{2} \left[ \log \sigma^2_B \cdot  1+ \frac{1}{\sigma^2_B} \cdot \sigma^2_B - (\mu^2_B+\sigma^2_B)\right]\\
    &=-\frac{1}{2} \left( \log \sigma^2_B + 1 - \mu^2_B-\sigma^2_B\right)
\end{aligned}
\end{equation}

When each dimension $d\in \{0,...,C\}$ is independent in multivariate Gaussian distribution $\mathcal{N}(\mu, \varepsilon)$, we have:
\begin{equation}
\begin{aligned}
    KL( p_\theta(B|S)\parallel q(B)) &= \int p_\theta(b|S) \log \frac{p_\theta(b_1|S) \cdots p_\theta(b_d|S)}{q(b_1) \cdots q(b_d)} \, db \\
    &= \int p_\theta(b|S) \left[ \sum_{d=1}^C \log p_\theta(b_d|S) - \sum_{d=1}^C \log q(b_d) \right] db \\
    &= \sum_{d=1}^C \int p_\theta(b|S) \left[ \log p_\theta(b_d|S) - \log q(b_d) \right] db \\
    &= \sum_{d=1}^C \int p_\theta(b_d|S) \left[ \log p_\theta(b_d|S) - \log q(b_d) \right] db_d \\
    &= \sum_{d=1}^C KL\left(p_\theta(b_d|S) \, \| \, q(b_d)\right)
\end{aligned}
\end{equation}
Then we have:
\begin{equation}
    \min KL( \mathcal{N}(\mu_B, \sigma^2_B)\parallel\mathcal{N}(0,\textbf{I})) =\min -\frac{1}{2} \mathbb{E}_C  \left(\log \sigma^2_B + 1 - \mu^2_B-\sigma^2_B\right)
\label{equ_ib_derive1}
\end{equation}
where $C$ is the feature dimension of the bottleneck latents $B$.

While for maximizing $\mathbb{E}_{B\sim p(B|S)}\mathbb{E}_{S\sim p(S)} [\log q_\phi(T|B)]$, we leverage diverse expression to conduct computation based on the property of specific target state $T$, divided into token- and label-level IB in this paper. 

\subsection{Derivation of Token-level Information Bottleneck} For unimodal representation $F_u=\{f_u\}|_{i=1}^L\in\mathbb{R}^{L\times C}$ as target state $T$, we utilize IB decoder $D_T:B\mapsto T$ to enable information flow across the source and target representations $F_S/F_T$ with a bottleneck bridge $B$. Since each dimension is independent for token embeddings of $F_S/F_T$, we have:
\begin{equation}
    q_\phi(T|B) = \mathcal{N}(\mu_T, \sigma^2_T) =\prod_{i=1}^{L}\ \prod_{d=1}^{C} \mathcal{N}(\mu^{d}_T, (\sigma_T^{d})^2) =\prod_{i=1}^{L}\ \left(\prod_{d=1}^{C}\frac{1}{\sqrt{2\pi}\sigma^d_T}\right)\exp{\left[-\sum_{i=1}^{C}\frac{(f^d_T-\mu^d_T)^2}{2(\sigma_T^{d})^2} \right]}
\end{equation}
Then we have:
\begin{equation}
    \log q_\phi(T|B) = -\sum^L_{i=1}\left[\frac{C}{2} \log{2\pi} + \frac{1}{2} \sum^C_{d=1} \log (\sigma_T^{d})^2 + \frac{1}{2} \sum^C_{d=1}  \frac{(f^d_T-\mu^d_T)^2}{(\sigma_T^{d})^2}\right]
\end{equation}

For simplify, assuming that variance $(\sigma_T^{d})^2$ of each dimension $d\in\{0,..,C\}$ is consistent and fixed as a constant, IB decoder only needs to output the mean $\mu^i=D_T(b_S)$ of the projected token embedding. Then the following equation holds:
\begin{equation}
\begin{aligned}
    \max \mathbb{E}_{B\sim p(B|S)}\mathbb{E}_{S\sim p(S)} [\log q_\phi(T|B)] &\equiv \min \sum^L_{i=1}\left[\frac{1}{2} \sum^C_{d=1} (f^d_T-\mu^d_T)^2\right] \\
    &\equiv \min \sum^L_{i=1}\mathbb{E}_{b_S} \parallel f^d_T-D_T(b_S)\parallel^2
\end{aligned}
\label{equ_tib_derive1}
\end{equation}

Combining Equ. \ref{equ_ib_derive1} and Equ. \ref{equ_tib_derive1}, we can derive Equ. \ref{equ_token_ib2} as the objective of $S\rightarrow T$ token-level information bottleneck:
\begin{equation}
\begin{aligned}
    \mathcal{L}^{S\rightarrow T}_{tib} &\approx \mathbb{E}_{F_S\sim p(F_S)} \left[KL( p_\theta(B_S|F_S)\parallel q(B_S))\right] - \beta\ \mathbb{E}_{B_S\sim p(B_S|F_S)}\mathbb{E}_{F_S\sim p(F_S)} [\log q_\phi(F_T|B_S)]\\
    &=\frac{1}{L}\sum^L_{i} \{KL( \mathcal{N}(\mu^i_B, (\sigma_B^{i})^2)\parallel\mathcal{N}(0,\textbf{I})) + \beta\ \mathbb{E}_{b_S} [\parallel f_T-D_T(b_S)\parallel^2] \} \\
    &= \frac{1}{L}\sum^L_{i} \{\left[-\frac{1}{2} \mathbb{E}_C  \left(\log \sigma^2_B + 1 - \mu^2_B-\sigma^2_B\right) \right]+ \beta\ \mathbb{E}_{b_S} [\parallel f_T-D_T(b_S)\parallel^2]\}
\end{aligned}
\end{equation}

\subsection{Derivation of Label-level Information Bottleneck} For ground truth label $y_{gt}\in\mathbb{R}^{1}/\mathbb{R}^{V}$ as target state $T$, we utilize two forms of the posterior distribution $q_\phi(T|B)$ according to the continuous ($1$ as the regression value) or discrete ($V$ as the one-hot recognition classes) forms of the supervision. We leverage unimodal predictor $P_S:B_S\mapsto \hat{y}_S$ for each source modality $S$ to model the posterior distribution $q_\phi(T|B)$ and output the prediction $\hat{y}_S$ based on information of single modality.

With continuous labels $y_{gt}\in\mathbb{R}^{1}$ in regression task, we consider $q_\phi(T|B)$ as one-dimensional Gaussian Distribution, computed as:
\begin{equation}
\begin{aligned}
    q_\phi(T|B) &= e^{-\parallel y_{gt}-\hat{y}_S\parallel} = e^{-\parallel y_{gt}-P_S(B_S)\parallel}\\
    \log{q_\phi(T|B)}&= \parallel y_{gt}-P_S(B_S)\parallel
\end{aligned}
\end{equation}

Then we turn the maximization of posterior distribution $q_\phi(T|B)$ into:
\begin{equation}
    \max \mathbb{E}_{B\sim p(B|S)}\mathbb{E}_{S\sim p(S)} [\log q_\phi(T|B)] \equiv \min \sum^L_{i=1}\mathbb{E}_{B_S} \parallel y_{gt}-P_S(B_S)\parallel
\label{equ_lib_derive1}
\end{equation}

Combining Equ. \ref{equ_ib_derive1} and Equ. \ref{equ_lib_derive1}, we can derive Equ. \ref{equ_label_ib_final1} as the objective of label-level information bottleneck:
\begin{equation}
    \mathcal{L}_{lib} \approx \frac{1}{N}\sum^N_{i} \left[-\frac{1}{2} \mathbb{E}_C  \left(\log \sigma^2_B + 1 - \mu^2_B-\sigma^2_B\right) \right] + \beta\ \mathbb{E}_{B_S} [\parallel y_{gt}-P_S(B_S)\parallel]
\end{equation}

With one-hot discrete labels $y_{gt}\in\mathbb{R}^{V}$ with $V$ classes in classification task,  we consider $q_\phi(T|B)$ as multi-dimensional Bernoulli Distribution, computed as:
\begin{equation}
\begin{aligned}
    q_\phi(T|B) &= \prod^V \hat{y}_S^{y_{gt}}= \prod^V [P_S(B_S)]^{y_{gt}}\\
    \log{q_\phi(T|B)}&= \sum^V y_{gt}\log{P_S(B_S)}
\end{aligned}
\end{equation}
Then we turn the maximization of posterior distribution $q_\phi(T|B)$ into:
\begin{equation}
    \max \mathbb{E}_{B\sim p(B|S)}\mathbb{E}_{S\sim p(S)} [\log q_\phi(T|B)] \equiv \min \mathbb{E}_{B_S} [\sum^V y_{gt}\log{P_S(B_S)}]
\label{equ_lib_derive2}
\end{equation}

Combining Equ. \ref{equ_ib_derive1} and Equ. \ref{equ_lib_derive2}, we can derive Equ. \ref{equ_label_ib_final2} as the objective of label-level information bottleneck:
\begin{equation}
    \mathcal{L}_{lib} \approx \frac{1}{N}\sum^N_{i} \left[-\frac{1}{2} \mathbb{E}_C  \left(\log \sigma^2_B + 1 - \mu^2_B-\sigma^2_B\right) \right] + \beta\ \mathbb{E}_{B_S} [\sum^V y_{gt}\log{P_S(B_S)}]
\end{equation}

\subsection{Reparameterization Trick for Informative Bottleneck Latent} 
Since the informative bottleneck latents $B$ is sampled from $p_\theta(B|S)\sim\mathcal{N}(\mu, \sigma^2)$ where the sampling process is not differentiable. For the purpose of optimizing the IB encoder $E_S:S\mapsto B$ through gradient decent algorithm, we need to identify the gradient of $B$ to the weight parameters $\theta$ of the encoder, denoted as follows according to the Chain Role:
\begin{equation}
    \frac{\partial B}{\partial\theta}=\frac{\partial B}{\partial\mu}\frac{\partial \mu}{\partial\theta} + \frac{\partial B}{\partial\sigma^2}\frac{\partial \sigma^2}{\partial\theta} 
\end{equation}
where $\mu$ and $\sigma$ are directly output by the encoder so that $\frac{\partial \mu}{\partial\theta}$ and $\frac{\partial \sigma^2}{\partial\theta}$ is easily tractable. 

To compute $\frac{\partial B}{\partial\mu}$ and $\frac{\partial B}{\partial\sigma^2}$ from the sampled bottleneck latent $B$, we firstly sample a random vector $\textbf{z}$ from nonparametric distribution $\mathcal{N}(0, \textbf{I})$ and utilize a transformation function $g(\textbf{z},\mu,\sigma^2)$ to attain the bottleneck $B$, shown as:
\begin{equation}
    B=\mu+\sigma\ \odot\ \textbf{z} \ \text{, where } \textbf{z} \sim \mathcal{N}(0,\textbf{I})
\end{equation}

Since the process of sampling $\textbf{z}\sim\mathcal{N}(0, \textbf{I})$ is independent to parameters $\mu$ and $\sigma$, the derivation of $\frac{\partial B}{\partial\mu}$ and $\frac{\partial B}{\partial\sigma^2}$ are transformed into the derivation of $\frac{\partial g(\cdot)}{\partial\mu}$ and $\frac{\partial g(\cdot)}{\partial\sigma^2}$, which is differentiable.

% \section{Algorithm for CyIN with Complete and Incomplete Multimodal Learning}
% \subsection{Pseudo Code for Train}

% \subsection{Pseudo Code for Inference}

\section{Theoretical Grounding of combining IB with Cyclic Translation} 
We present further explanation for the theoretical analysis of CyIN in combining Information Bottleneck (IB) with cross-modal cyclic  translation. Except for task prediction term, the optimization objective for IB loss and cyclic translation in Equ. \ref{equ_loss_total} is deduced as follows:
\begin{equation}
\begin{aligned}
\mathcal{L} &= \frac{1}{\beta}(\mathcal{L}_{tib}+\mathcal{L}_{lib})+\gamma \mathcal{L}_{tran} \\
&= \frac{1}{\beta}\mathcal{L}_{IB}+\gamma (\mathcal{L}_{rec}+\mathcal{L}_{cyc})
\end{aligned}
\end{equation}

Considering multimodal learning with two modalities without loss of generality, the theoretical modeling of CyIN can be formulated as follows:
\begin{alignat*}{2}
    S_1 \rightarrow & B_1 \rightleftharpoons\  &&B_2  \leftarrow S_2 \\
    &\downarrow &&\downarrow \\
    &T_1 &&T_2
\end{alignat*}
where $S$ denotes the source state and $T$ denotes the target state. 

\begin{itemize}[leftmargin=*]

\item The left and right side with $S_1 \rightarrow B_1 \rightarrow T_1$ and $S_2 \rightarrow B_2 \rightarrow T_2$ denote the chain of information bottleneck. Note that the source and target states have $S/T\in\{F_{S_1},F_{S_2}\}\{F_{T_1},F_{T_2}\}$ in the token-level IB or $S\in\{F_{S_1},F_{S_2}\},T\in y$ in the label-level IB, both of which have been theoretically deduced above, formulated as:

\begin{equation}
\mathcal{L}_{tib} = \underbrace{[I(S_1;B_1)-\beta I(B_1;T_1)]}_{IB\ for\ modality\ 1} + \underbrace{[I(S_2;B_2)-\beta I(B_2;T_2)]}_{IB\ for\ modality\ 2}
\end{equation}

\item The middle part $B_1 \rightleftharpoons B_2$ denote the cyclic translation with $B_1 \rightharpoonup B^{rec}_{2\rightarrow 1}\rightharpoonup B^{cyc}_1 \sim B_1$ and $B_2 \rightharpoonup B^{rec}_{1\rightarrow2}\rightharpoonup B^{cyc}_2 \sim B_1$, where $\rightharpoonup$ denotes the translation process while $\sim$ denotes alignment with the original bottleneck. Considering $\Gamma$ as the translator network, the translation process can be divided into cross-modal reconstruction across diverse unimodal latents $B_1/B_2$ and cyclic translation from the reconstructed latent $B^{rec}_{1\rightarrow2}/B^{rec}_{2\rightarrow1}=\Gamma_{1\rightarrow 2}(B_1)/\Gamma_{2\rightarrow 1}(B_2)$ back to the origin $B^{cyc}_2/B^{cyc}_1=\Gamma_{1\rightarrow 2} (B^{rec}_{2\rightarrow 1})/\Gamma_{2\rightarrow 1} (B^{rec}_{1\rightarrow 2})$, formulated as:
\begin{equation}
\begin{aligned}
\mathcal{L}_{rec}&= \underbrace{D_{KL}(B_1;B^{rec}_{2\rightarrow 1})]}_{Reconstruction\ from\ modality\ 1 \rightharpoonup\ 2}+\underbrace{D_{KL}(B_2;B^{rec}_{1\rightarrow 2})}_{Reconstruction\ from\ modality\ 2 \rightharpoonup\ 1} \\
\mathcal{L}_{cyc}&=\underbrace{D_{KL}(B_2;B^{cyc}_2)}_{Translating\ Back\ from\ modality\ 2 \rightharpoonup\ 1} +\underbrace{{D_{KL}(B_1;B^{cyc}_1)}}_{Translating\ Back\ from\ modality\ 1 \rightharpoonup\ 2}
\end{aligned}
\end{equation}

\end{itemize}

Minimizing above losses, we can obtain informative unimodal bottleneck latents $B_u$ for each modality $u$, which contains inter-modal features to conduct productive multimodal fusion. Comparing with directly fusing unimodal representations $F_u$, fusion process implemented on the bottleneck latents $B_u$, denoted as $F_M=D_M(B_1,B_2)$, can benefit from the compression ability of information bottleneck and lead to less reconstruction difficulty in cross-modal translation.

When both modalities $u_1$ and $u_2$ are presented, known as complete multimodal learning, we can regard the cyclic translation $B_1 \rightleftharpoons B_2$ as sort of cross-modal interaction enhancing the extraction of modality-shared dynamics. Such cross-modal synergies can benefit the multimodal fusion process $F_M=D_M(B_1,B_2)$ to attain the final discriminative representation. 

When one of the modalities is missing, the framework is required to conduct incomplete multimodal learning. Assuming $u_1$ is missing, we can obtain bottleneck latents $B_2$ from modality $u_2$ to reconstruct the bottleneck latents $B^{rec}_1$, denoted as $B_2 \rightharpoonup B^{rec}_1$, which can be further decoded as the supplementary information for the multimodal fusion process $F_M=D_M(B^{rec}_1,B_2)$. 

The aforementioned two-modality scenarios can be generalized to arbitrary modality pairs without constraint, thereby facilitating efficient scaling of CyIN for tasks involving multiple modalities.

% Model Architecture
\section{Hyper-parameter Setting} 
\label{appendix_hyperparam}

Since missing modalities lead to highly performance fluctuations under different missing scenarios, to guarantee fair and consistent comparison among various methods, we conduct evaluation of each experiment ten times using fixed 10 random seeds. We utilize AdamW \cite{loshchilov2018decoupled} as the training optimizer. %The division rate of the first and second stage is set as $0.2$ for better convergence, meaning 5:4 training time for jointly complete and incomplete multimodal learning.

The split of datasets and hyper-parameters of CyIN are reported in Table \ref{table_dataset_hyperparameter}.

\begin{table}[htbp]
\caption{Splits of datasets and the corrsponding hyper-parameters settings. }\label{table_dataset_hyperparameter}
% Note that 'LR' denotes the learning rate of trainable parameters.
\centering
\scalebox{0.9}{
\begin{tabular}{ccccc}
\toprule[1pt]
    Multimodal Task & \multicolumn{2}{c}{Regression} & \multicolumn{2}{c}{Classification} \\ % Table header row
\cmidrule(r){2-3}\cmidrule(r){4-5}
  Dataset & MOSI & MOSEI & IEMOCAP & MELD \\ 
\midrule[1pt]
  Train & 1,284 & 16,326 & 5,228 & 9,765 \\
  Valid & 229 & 1,871 & 519 & 1,102 \\
  Test & 686 & 4,659 & 1,622 & 2,524 \\
\midrule[1pt]
  Training Epochs & 50  &  30 &  50 &  50 \\
  Batch Size &  128 &  128 &  256 & 128  \\
  Learning Rate of Language Model & 4e-5  &  1e-5 &  5e-6 & 3e-5  \\
  Learning Rate of Other Parameters & 1e-3  &  5e-4 &  1e-4 & 5e-4  \\
  Weight Decay & 1e-2  &  1e-5 & 1e-3  &  1e-4 \\
  Dimension $C_U$ for Unimodal Representation &  256 & 64  & 64 & 32  \\
  Dimension $C_{ib}$ in IB Encoder or Decoder & 256  &  64 & 256  &  64 \\
  Dimension $C_B$ of Bottleneck Latent $B$ &  128 &  256 & 128  &  32 \\
  \# Layers of $RA(\cdot)$ in CRA & 8  & 16  & 8  & 4  \\
  Dimension of $RA(\cdot)$ in CRA & $\left[64,32,16\right]$  & $\left[128,64,32\right]$  & $\left[128,64,32\right]$  & $\left[128,64,32\right]$  \\
  \# Layers of Cross-modal Attention & 2  & 8  &  2 &  4 \\
  \# Heads of Attention $H$ & 8  &  2 &  4 & 8  \\
  $\beta$ &  16 &  32 &  4 &  4 \\
  $\gamma$ &  10 &  5 &  10 & 5  \\
  $1^{st}$:$2^{nd}$ Training Stage  & 1:9 & 3:7 & 3:7 & 1:4 \\
\bottomrule[1pt]
\end{tabular}
}
% \vspace{-0.2cm}
\end{table}

\section{Details about Evaluation Protocols} \label{appendix_miss_protocol}
The evaluation under the circumstance of (1) complete multimodal input is consistent with other state-of-the-art baselines, referring to $u \in \{l,a,v\}$, where $l$, $a$, and $v$ denote language, acoustic, and visual modalities, respectively.

Following \cite{zhao2021missing,lian2023gcnet, wang2023incomplete,zhang2024towards}, to evaluate model robustness under missing modality scenarios, we adopt two missing data protocols: (2) fixed and (3) random missing protocols. 

(2) In the fixed missing protocol, a consistent subset of modalities is removed across all iterations. Specifically, we discard either one modality (e.g., $u \in \{l,a\}/\{l,v\}/\{a,v\}$) or two modalities (e.g., $u \in \{l\}/\{a\}/\{v\}$). 

(3) In contrast, the random missing protocol simulate the real-world scenarios by randomly selecting missing modality combinations for each sample in every batch, where either one or two modalities may be absent. Following \cite{lian2023gcnet,wang2023incomplete}, the random missing rate ($MR$) is defined as:
\begin{equation}
    MR=1-\frac{\sum^N_{i=1}|u_i|}{N\times U}
\end{equation}
where $|u_i|$ denotes the number of available modalities for the $i$-th sample, $U$ is the total number of modalities, and $N$ is the total number of samples. Given a realistic application scenario, each sample is guaranteed to have at least one available modality, ensuring $|u_i| \in \{1,...,U\}$, and consequently, $MR \leq \frac{U-1}{U}$. In our multimodal regression and classification tasks with $U=3$ modalities, we consider missing rates $MR \in \{0.0, 0.1, \dots, 0.7\}$, where $MR=0.7$ approximates the maximum tolerable missingness. Notably, $MR=0.0$ represents the complete multimodal scenario where all modalities ($u \in {l,a,v}$) are present.
 
\section{Baselines} \label{appendix_baseline}
Baselines are reproduced by the open-source codes. Following \cite{gkoumas2021makes}, we conduct fifty-times of random grid search for the best hyper-parameters of each model. The descriptions of baselines are presented as follows.

CCA \cite{hotelling1992cca} projects paired modalities into a shared low-dimensional space by maximizing their canonical correlations, providing a classical linear imputation for incomplete data.

DCCA \cite{andrew2013dcca} replaces the linear projections with deep neural encoders, enabling nonlinear correspondences between modalities.

DCCAE \cite{wang2015dccae} augments DCCA with autoencoder-based reconstruction objectives \cite{bengio2006greedy}, jointly preserving each modality’s internal structure while learning maximally canonical correlations.

CRA \cite{tran2017cra} employs a cascade of residual autoencoders that iteratively refine the reconstruction of representations with complete inputs from the ones with partial inputs.

MCTN \cite{pham2019found} leverages encoder–decoder recurrent neural networks to translate between source and target modalities in cyclic translation way.

MMIN \cite{zhao2021missing} extends cycle consistency learning with a cascaded residual architecture by imagining missing information from the observed views.

CPM-Net \cite{zhang2022cpmnet} utilize partial multi-view clustering to tackle incomplete multimodal learning issue, embedding all views into a structured latent space where missing features can be inferred via data transmission and cluster-aware classification.

GCNet \cite{lian2023gcnet} introduces two complementary graph neural networks to model temporal and speaker relationships in conversational data, and then reconstructs missing features from the learned graph representations.

IMDer \cite{wang2023incomplete} adopts score-based diffusion models to learn distribution of missing modalities in iterative diffusion and denoising process, regularized by the remained modalities.

LNLN \cite{zhang2024towards} employs a dominant modality correction module to ensure the quality of dominant modality representations and conduct dominant modality based multimodal learning to resist input noise.

\section{Evaluation Metrics}  \label{appendix_metric}
The formulas and computation for evaluation metrics on multimodal regression and classification tasks are presented as follows.

For multimodal regression tasks on MOSI and MOSEI, we adopt well-representative metrics including \textbf{Acc7, F1, MAE, and Corr}, to evaluate models' performance \cite{poria2017review}:

\textbf{Acc7 (Seven‐class Accuracy)} measures the proportion of correct predictions across the seven integer labels in $\left[-3, +3\right]$, formulated as:
\begin{equation}
\mathrm{Acc7}
= \frac{1}{N}\sum_{n=1}^{N}
\mathbb{I}\left( \lfloor \hat y_n \rceil = \lfloor y_n \rceil) \right)
=\frac{1}{7}
\sum_{k=-3}^{3}
\frac{\bigl|\{\,n : y_n = k \land \hat y_n = k\}\bigr|}
     {\bigl|\{\,n : y_n = k\}\bigr|}
\end{equation}
where $N$ is the total number of samples, $\hat y_n$ is the predicted label, $y_n$ is the ground‐truth label, $\lfloor \cdot \rceil$ denotes the round to nearest integer and \(\mathbb{I}(\cdot)\) is the indicator function. With $k\in\{-3, -2, -1, 0, 1, 2, 3\}$, $\{n : y_n = k\}$ denotes the set of samples with true label $k$, and $\{n : y_n = k \land \hat y_n = k\}$ is the subset correctly predicted as $k$.

\textbf{F1 (Binary F1‐score)} assesses positive–negative discrimination by collapsing regression labels into two classes (negative/positive), formulated as:
\begin{equation}
F1=\frac{1}{N}\sum_{n=1}^{N} w_c\cdot 2 \cdot \frac{(Precision_c * Recall_c)} {(Precision_c + Recall_c)}
\label{equ_f1}
\end{equation}
\begin{equation}
Precision=\frac{TP}{TP + FP}\text{ ,  } Recall = \frac{TP}{TP + FN}
\end{equation}
where $C$ is the number of classes, $w_c$ is the weight of class $c$ (typically $w_c = N_c / N$), and $TP_c$, $FP_c$, and $FN_c$ denote the numbers of true positives, false positives, and false negatives for class $c$, respectively.

\textbf{MAE (Mean Absolute Error)} quantifies the average absolute deviation between predicted and ground‐truth  scores, formulated as:
\begin{equation}
MAE(\hat{y}, y) = \frac{ {\sum_{n=1}^{N}}\left | \hat{y}_n-y_n \right |  }{N}
\end{equation}

\textbf{Corr (Pearson Correlation)} captures the linear agreement between predictions and labels, indicating any systematic bias or skew in model outputs, formulated as:
\begin{equation}
\text{Corr}(\hat{y}, y)=\frac{ {\textstyle \sum_{n=1}^{N}}\left ( \hat{y}_n-\bar{\hat{y}} \right )\left ( y_n-\bar{y} \right )  }{\sqrt{ {\textstyle \sum_{n=1}^{N}}\left ( \hat{y}_n-\bar{\hat{y}} \right )^2{\textstyle \sum_{n=1}^{N}}\left ( y_n-\bar{y} \right )^2  } }
\end{equation}
where $\bar{\hat{y}}$ and $\bar{y}$ are the means of the predicted scores and ground truth labels.

For multimodal classification tasks on IEMOCAP and MELD, we report \textbf{Acc} and \textbf{wF1} to balance the weight of scores from each class \cite{geetha2024multimodal}.

\textbf{Acc (Binary Accuracy)} describes the unweighted proportion of correctly predicted samples out of the total number of samples, formulated as:
\begin{equation}
Acc=\frac{TP+TN}{TP +TN + FP + FN}
\end{equation}

\textbf{wF1 (Weighted F1‐score)} balances weighted precision and recall based on the true instances for each class, and the formulation of wF1 in classification is the same as Equ. \ref{equ_f1}.

\section{More Experiment Results} 
\label{appendix_more_result}

% \subsection{Loss Tracing}
% \label{appendix_loss_trace}

\subsection{Detailed Results Under Various Missing Scenarios}
\label{appendix_details_result}

\begin{table}[htbp]
\centering
\setlength\tabcolsep{7.5pt}
\caption{Performance comparison between the proposed CyIN and baselines with fixed missing protocols including incomplete settings $u\in\{l\}/\{v\}/\{a\}/\{l,a\}/\{l,v\}/\{a,v\}$. }
\label{table_fixmiss_detail}
\scalebox{0.68}{\begin{tabular}{cccccccccccccc}
    \toprule[1.5pt]
    \multirow{2}{*}{Fix $u$} & \multirow{2}{*}{Models} & \multicolumn{4}{c}{MOSI} & \multicolumn{4}{c}{MOSEI} & \multicolumn{2}{c}{IEMOCAP} & \multicolumn{2}{c}{MELD}\\
    \cmidrule(r){3-6}\cmidrule(r){7-10}\cmidrule(r){11-12}\cmidrule(r){13-14}
    & & Acc7$\uparrow$ & F1$\uparrow$ & MAE$\downarrow$ & Corr$\uparrow$ & Acc7$\uparrow$ & F1$\uparrow$ & MAE$\downarrow$ & Corr$\uparrow$ & Acc$\uparrow$ & wF1$\uparrow$ & Acc$\uparrow$ & wF1$\uparrow$\\

    \midrule[0.5pt]
     \multirow{11}{*}{$\{l\}$} ~ & CCA & 26.4  & 74.7  & 1.098  & 0.567  & 44.6  & 80.2  & 0.677  & 0.621  & 56.1  & 53.9  & 49.7  & 43.1  \\ 
        ~ & DCCA & 23.3  & 73.5  & 1.196  & 0.520  & 41.5  & 68.0  & 0.810  & 0.377  & 51.7  & 50.7  & 48.9  & 35.3  \\ 
        ~ & DCCAE & 25.8  & 66.4  & 1.278  & 0.508  & 41.6  & 67.9  & 0.806  & 0.336  & 50.1  & 49.2  & 47.7  & 32.0  \\ 
        ~ & CPM-Net & 17.2  & 63.9  & 1.335  & 0.345  & 40.3  & 74.7  & 0.832  & 0.230  & 50.4  & 50.5  & 42.0  & 34.8  \\ 
        ~ & CRA & 36.3  & 82.8  & 0.900  & 0.749  & 49.3  & 84.9  & 0.576  & 0.741  & 53.8  & 51.3  & 55.3  & 51.6  \\ 
        ~ & MCTN & 41.5  & 83.7  & 0.777  & 0.765  & 49.6  & 81.8  & 0.594  & 0.715  & 58.4  & 57.8  & 56.1  & 52.3  \\ 
        ~ & MMIN & 42.3  & 85.0  & 0.743  & 0.791  & \textbf{52.8}  & 84.6  & 0.543  & \textbf{0.761}  & 57.0  & 57.5  & 59.2  & 54.9 \\ 
        ~ & GCNet & 42.6  & 85.0  & 0.740  & 0.795  & 51.0  & 84.8  & 0.562  & 0.750  & 58.7  & 58.3  & 59.9  & 57.9  \\ 
        ~ & IMDer & 43.0  & 85.1  & \textbf{0.717}  & 0.794  & 51.8  & 85.0  & 0.552  & 0.756  & 60.4  & 60.1  & 59.9  & 58.2  \\ 
        ~ & LNLN & 43.9  & 83.5  & 0.761  & 0.760  & 52.5  & 84.3  & 0.547  & 0.765  & 62.8  & 62.4  & 58.3  & 57.1  \\ 
        ~ & \textbf{CyIN} & \textbf{44.0} & \textbf{85.2} & 0.742 & \textbf{0.795} & 52.6 & \textbf{85.4} & \textbf{0.541} & 0.751 & \textbf{63.4} & \textbf{63.0} & \textbf{60.0} & \textbf{58.6} \\ 
    \midrule[0.5pt]
    \multirow{11}{*}{$\{a\}$} ~ & CCA & 17.5  & 46.8  & 1.422  & 0.112  & 41.5  & 63.3  & 0.808  & 0.272  & 39.1  & 34.8  & 47.7  & 35.7  \\ 
        ~ & DCCA & 14.1  & 43.3  & 1.821  & 0.052  & 41.3  & 59.4  & 0.810  & 0.356  & 30.0  & 24.8  & 46.1  & 33.4  \\ 
        ~ & DCCAE & 19.7  & 56.5  & 1.406  & 0.123  & 37.4  & 50.8  & 0.841  & 0.301  & 35.6  & 34.0  & 47.3  & 34.4  \\ 
        ~ & CPM-Net & 17.3  & 56.6  & 1.371  & 0.313  & 39.4  & \textbf{74.2}  & 0.916  & 0.139  & 38.1  & 37.2  & 26.3  & 29.7  \\ 
        ~ & CRA & 19.7  & 52.0  & 1.438  & 0.102  & 41.8  & 71.1  & \textbf{0.781}  & \textbf{0.420}  & 40.7  & 40.1  & 46.6  & 37.5  \\ 
        ~ & MCTN & 15.5  & 42.2  & 1.431  & 0.018  & 41.4  & 62.9  & 0.842  & 0.096  & 22.3  & 11.3  & 48.2  & 31.6  \\ 
        ~ & MMIN & 20.1  & 52.6  & 1.429  & 0.080  & 39.3  & 70.9  & 0.790  & 0.404  & 42.6  & 40.0  & 48.2  & 31.5 \\ 
        ~ & GCNet & 19.8  & 57.5  & 1.383  & 0.387  & 40.8  & 56.6  & 0.817  & 0.389  & 44.6  & 42.2  & 40.9  & 34.0  \\ 
        ~ & IMDer & 19.6  & 58.3  & 1.363  & 0.392  & 42.1  & 69.9  & 0.795  & 0.372  & 47.0  & 46.8  & 45.6  & \textbf{42.5} \\ 
        ~ & LNLN & 15.6  & 47.7  & 1.442  & 0.075  & 41.0  & 68.9  & 0.795  & 0.338  & 39.2  & 33.4  & 39.3  & 33.7  \\ 
    & \textbf{CyIN}  & \textbf{20.7} & \textbf{60.2} & \textbf{1.355} & \textbf{0.412} & \textbf{42.3} & 70.8 & 0.782 & 0.363 & \textbf{50.3} & \textbf{48.9} & \textbf{48.5} & 37.7 \\ 
    \midrule[0.5pt]
    \multirow{11}{*}{$\{v\}$} ~ & CCA & 15.5  & 26.1  & 1.445  & 0.125  & 41.4  & 48.6  & 0.837  & 0.239  & 23.6  & 9.1  & \textbf{48.4}  & 31.6  \\ 
        ~ & DCCA & 17.4  & 51.7  & 1.437  & 0.081  & 41.3  & 72.9  & 0.800  & 0.387  & 26.6  & 22.4  & 45.8  & 32.9  \\ 
        ~ & DCCAE & \textbf{20.9}  & \textbf{61.8}  & 1.659  & 0.407  & 41.4  & 73.1  & 0.798  & 0.389  & 25.2  & 21.8  & 45.5  & 32.9  \\ 
        ~ & CPM-Net & 16.5  & 56.8  & 1.368  & 0.323  & 38.8  & \textbf{74.2}  & 0.862  & 0.205  & 28.0  & 20.9  & 31.8  & 32.1  \\ 
        ~ & CRA & 16.6  & 52.1  & 1.396  & 0.013  & 39.0  & 72.0  & 0.781  & 0.426  & 29.9  & 26.3  & 44.2  & 36.8  \\ 
        ~ & MCTN & 15.5  & 42.2  & 1.431  & 0.018  & 41.4  & 62.9  & 0.842  & 0.096  & 23.5  & 8.9  & \textbf{48.4}  & 31.6  \\ 
        ~ & MMIN & 19.7  & 50.5  & 1.471  & 0.075  & 39.5  & 69.8  & 0.789  & 0.395  & 35.3  & 34.1  & \textbf{48.4}  & 31.6 \\ 
        ~ & GCNet & 16.0  & 47.9  & 1.390  & 0.066  & 41.4  & 58.5  & 0.833  & 0.268  & 44.8  & 43.4  & 39.1  & 33.5  \\ 
        ~ & IMDer & 17.0  & 52.0  & 1.426  & 0.032  & 41.4  & 62.7  & 0.826  & 0.386  & 46.8  & 45.8  & 46.3  & 38.6 \\ 
        ~ & LNLN & 15.5  & 42.2  & 1.447  & 0.125  & 41.8  & 70.9  & 0.771  & 0.405  & 47.9  & 47.3  & 48.1  & 31.3  \\ 
    & \textbf{CyIN}  & 19.8 & 57.7 & \textbf{1.362} & \textbf{0.408} & \textbf{42.1} & 71.7 & \textbf{0.767} & \textbf{0.430} & \textbf{49.0} & \textbf{47.7} & 48.2 & \textbf{39.3} \\ 
    \midrule[0.5pt]
    \multirow{11}{*}{$\{l,a\}$} ~ & CCA & 27.8  & 74.9  & 1.106  & 0.542  & 44.5  & 81.8  & 0.663  & 0.639  & 60.1  & 58.5  & 51.1  & 46.5  \\ 
        ~ & DCCA & 21.9  & 59.7  & 1.377  & 0.422  & 41.3  & 72.8  & 0.795  & 0.401  & 53.9  & 52.5  & 47.9  & 35.9  \\ 
        ~ & DCCAE & 22.2  & 69.4  & 1.425  & 0.414  & 38.7  & 68.2  & 0.802  & 0.394  & 54.0  & 52.7  & 47.3  & 34.5  \\ 
        ~ & CPM-Net & 17.2  & 64.0  & 1.335  & 0.345  & 39.9  & 74.5  & 0.980  & 0.145  & 43.9  & 44.0  & 35.1  & 35.8  \\ 
        ~ & CRA & 36.4  & 83.4  & 0.885  & 0.750  & 50.5  & 84.6  & 0.565  & 0.756  & 60.7  & 60.0  & 56.4  & 53.3  \\ 
        ~ & MCTN & 41.5  & 83.7  & 0.777  & 0.765  & 47.9  & 84.3  & 0.592  & 0.721  & 58.4  & 57.8  & 56.3  & 52.4  \\ 
        ~ & MMIN & 43.0  & 85.0  & 0.744  & 0.782  & 52.8  & 84.6  & 0.538  & 0.768  & 62.2  & 62.5  & 59.4  & 55.4 \\ 
        ~ & GCNet & 41.5  & 84.8  & 0.742  & 0.796  & 49.2  & 84.4  & 0.581  & 0.753  & 62.9  & 63.0  & 57.4  & 55.8  \\ 
        ~ & IMDer & 44.3  & 84.9  & 0.731  & \textbf{0.797}  & 52.8  & 85.1  & 0.560  & 0.775  & 63.7  & 63.9  & 60.0  & 58.3 \\ 
         ~ & LNLN & 44.0  & 83.3  & 0.761  & 0.766  & 52.7  & 84.6  & 0.546  & 0.768  & 57.9  & 57.3  & 58.1  & 57.1  \\ 
    & \textbf{CyIN}  & \textbf{45.0} & \textbf{85.1} & \textbf{0.728} & 0.792 & \textbf{53.1} & \textbf{85.5} & \textbf{0.541} & \textbf{0.786} & \textbf{65.2} & \textbf{64.7} & \textbf{60.5} & \textbf{58.3} \\ 
    \midrule[0.5pt]
    \multirow{11}{*}{$\{l,v\}$} ~ & CCA & 26.1  & 75.6  & 1.093  & 0.570  & 44.7  & 80.7  & 0.674  & 0.625  & 56.6  & 54.5  & 49.7  & 43.1  \\ 
        ~ & DCCA & 24.1  & 71.2  & 1.234  & 0.398  & 41.2  & 73.3  & 0.791  & 0.420  & 50.0  & 49.9  & 47.2  & 35.3  \\ 
        ~ & DCCAE & 21.7  & 67.3  & 1.327  & 0.340  & 39.7  & 73.1  & 0.791  & 0.394  & 52.9  & 50.4  & 46.8  & 34.1  \\ 
        ~ & CPM-Net & 17.5  & 62.9  & 1.339  & 0.333  & 38.1  & 73.4  & 1.582  & 0.072  & 42.8  & 43.5  & 41.7  & 35.2  \\ 
        ~ & CRA & 36.3  & 81.5  & 0.917  & 0.741  & 51.3  & 85.1  & 0.555  & 0.761  & 58.3  & 55.7  & 56.4  & 53.3  \\ 
        ~ & MCTN & 41.5  & 83.7  & 0.777  & 0.765  & 47.9  & 84.3  & 0.592  & 0.721  & 58.3  & 58.2  & 56.3  & 52.4  \\ 
        ~ & MMIN & 42.3  & 85.0  & 0.742  & 0.791  & 52.8  & 85.1  & 0.534  & 0.768  & 59.3  & 59.4  & 60.1  & 55.5 \\ 
        ~ & GCNet & 39.0  & 84.4  & 0.771  & 0.796  & 49.1  & 84.8  & 0.582  & 0.750  & 56.6  & 56.2  & 60.2  & 58.3  \\ 
        ~ & IMDer & 45.8  & 85.1  & \textbf{0.701}  & \textbf{0.799}  & 52.9  & 85.2  & 0.556  & 0.761  & 60.2  & 60.0  & 59.8  & 58.0 \\ 
        ~ & LNLN & 43.7  & 83.5  & 0.760  & 0.766  & 52.7  & 85.0  & 0.544  & 0.770  & 62.1  & 61.5  & 58.2  & 57.1  \\ 
    & \textbf{CyIN}  & \textbf{46.2} & \textbf{85.3} & 0.727 & 0.794 & \textbf{53.5} & \textbf{85.2} & \textbf{0.531} & \textbf{0.778} & \textbf{63.2} & \textbf{62.9} & \textbf{60.9} & \textbf{58.8} \\ 
    \midrule[0.5pt]
    \multirow{11}{*}{$\{a,v\}$}  ~ & CCA & 17.4  & 48.1  & 1.420  & 0.117  & 41.6  & 63.6  & 0.806  & 0.278  & 39.6  & 35.1  & 48.0  & 36.1  \\ 
        ~ & DCCA & 17.5  & 53.6  & 1.445  & 0.093  & 41.2  & 73.6  & 0.790  & 0.408  & 35.4  & 33.3  & 46.1  & 33.7  \\ 
        ~ & DCCAE & 19.4  & 55.2  & 1.569  & 0.046  & 37.9  & 72.7  & 0.796  & 0.404  & 35.5  & 33.2  & 46.5  & 35.7  \\ 
        ~ & CPM-Net & 16.8  & 56.2  & 1.367  & 0.330  & 35.2  & 72.0  & 1.395  & 0.041  & 44.5  & 42.4  & 25.6  & 28.9  \\ 
        ~ & CRA & 18.5  & 55.0  & 1.410  & 0.066  & 40.4  & \textbf{73.2}  & 0.773  & 0.453  & 44.1  & 42.0  & 45.3  & 40.1  \\ 
        ~ & MCTN & 15.5  & 42.2  & 1.431  & 0.018  & 41.4  & 62.9  & 0.842  & 0.102  & 22.7  & 11.5  & 48.8  & 31.6  \\ 
        ~ & MMIN & 20.3  & 52.3  & 1.427  & 0.081  & 39.8  & 71.0  & 0.774  & 0.431  & 48.2  & 47.4  & 48.3  & 31.5 \\ 
        ~ & GCNet & 17.9  & 57.2  & 1.363  & 0.385  & 41.2  & 72.2  & 0.805  & 0.393  & 49.0  & 48.1  & 44.7  & 40.9  \\ 
        ~ & IMDer & 18.4  & 58.1  & 1.319  & 0.386  & 41.5  & 71.6  & 0.788  & 0.446  & 50.0  & 49.9  & 46.7  & 43.0 \\ 
         ~ & LNLN & 15.6  & 48.7  & 1.441  & 0.076  & 40.8  & 71.7  & 0.775  & 0.441  & 51.5  & 50.4  & 35.5  & 31.8  \\ 
    & \textbf{CyIN}  & \textbf{21.0} & \textbf{59.4} & \textbf{1.305} & \textbf{0.391} & \textbf{41.7} & 72.9 & \textbf{0.773} & \textbf{0.458} & \textbf{53.0} & \textbf{52.1} & \textbf{48.4} & \textbf{43.5} \\ 
    % \midrule[0.5pt]
    % \multirow{11}{*}{Avg.} &  CCA & & & & & & & & & & & &\\
    % & DCCA & & & & & & & & & & & &\\
    % & DCCAE & & & & & & & & & & & &\\
    % & CPM-Net & & & & & & & & & & & &\\
    % & CRA & & & & & & & & & & & &\\
    % & MCTN & & & & & & & & & & & &\\
    % & MMIN & & & & & & & & & & & &\\
    % & GCNet & & & & & & & & & & & &\\
    % & IMDer & & & & & & & & & & & &\\
    % & \textbf{CyIN}  & & & & & & & & & & & &\\
    \bottomrule[1.5pt]
\end{tabular}
}
\end{table}

\begin{table}[htbp]
\centering
\setlength\tabcolsep{7.5pt}
\caption{Performance comparison between the proposed CyIN and baselines with random missing protocols including missing rates $MR\in[0.1,0.2,0.3,0.4,0.5,0.6,0.7]$. }
\label{table_randommiss_detail}
\scalebox{0.68}{
    \begin{tabular}{cccccccccccccc}
    \toprule[1.5pt]
    \multirow{2}{*}{$MR$} & \multirow{2}{*}{Models} & \multicolumn{4}{c}{MOSI} & \multicolumn{4}{c}{MOSEI} & \multicolumn{2}{c}{IEMOCAP} & \multicolumn{2}{c}{MELD}\\
    \cmidrule(r){3-6}\cmidrule(r){7-10}\cmidrule(r){11-12}\cmidrule(r){13-14}
    & & Acc7$\uparrow$ & F1$\uparrow$ & MAE$\downarrow$ & Corr$\uparrow$ & Acc7$\uparrow$ & F1$\uparrow$ & MAE$\downarrow$ & Corr$\uparrow$ & Acc$\uparrow$ & wF1$\uparrow$ & Acc$\uparrow$ & wF1$\uparrow$\\

    \midrule[1.5pt]
     \multirow{11}{*}{$0.1$} ~ & CCA & 26.9  & 72.7  & 1.133  & 0.512  & 45.6  & 81.2  & 0.671  & 0.635  & 58.9  & 58.7  & 49.3  & 41.9  \\ 
        ~ & DCCA & 24.1  & 70.1  & 1.219  & 0.418  & 40.1  & 73.1  & 0.789  & 0.415  & 52.4  & 51.4  & 47.5  & 35.3  \\ 
        ~ & DCCAE & 20.2  & 68.4  & 1.620  & 0.336  & 38.7  & 72.5  & 0.787  & 0.418  & 51.0  & 50.0  & 46.4  & 35.5  \\ 
        ~ & CPM-Net & 16.8  & 63.9  & 1.336  & 0.339  & 34.5  & 74.3  & 1.780  & 0.060  & 55.1  & 55.1  & 42.5  & 33.2  \\ 
        ~ & CRA & 33.6  & 79.3  & 0.972  & 0.699  & 50.1  & 84.2  & 0.576  & 0.736  & 60.0  & 59.0  & 56.0  & 53.0  \\ 
        ~ & MCTN & 38.9 & 79.9 & 0.846 & 0.720  & 47.1  & 81.4  & 0.620  & 0.680  & 42.0  & 43.3  & 53.6  & 46.8  \\ 
        ~ & MMIN & 41.0  & 82.1  & 0.808  & 0.741  & 51.5  & 83.5  & 0.562  & 0.738  & 53.9  & 53.9  & 55.8  & 49.0 \\ 
        ~ & GCNet & 41.2  & 84.5  & 0.806  & 0.751  & 50.3  & 83.1  & 0.653  & 0.696  & 60.7  & 60.8  & 56.3  & 54.4  \\ 
        ~ & IMDer & 42.3  & 84.1  & 0.796  & 0.753  & 51.4  & 83.9  & 0.600  & 0.721  & 60.9  & 61.3  & 58.4  & \textbf{56.7} \\ 
        ~ & LNLN & 40.8  & 83.3  & 0.790  & 0.756  & 51.4  & 83.7  & 0.566  & 0.745  & 61.2  & 62.0  & 56.2  & 55.0  \\ 
    & \textbf{CyIN} & \textbf{42.5} & \textbf{84.6} & \textbf{0.783} & \textbf{0.759} & \textbf{51.9} & \textbf{84.5} & \textbf{0.562} & \textbf{0.748} & \textbf{64.0} & \textbf{63.9} & \textbf{58.6} & 56.6 \\ 
    \midrule[0.5pt]
    \multirow{11}{*}{$0.2$} ~ & CCA & 25.6  & 70.4  & 1.161  & 0.489  & 45.0  & 79.3  & 0.688  & 0.605  & 56.0  & 55.8  & 49.1  & 41.3  \\ 
        ~ & DCCA & 22.8  & 68.7  & 1.242  & 0.394  & 40.4  & 72.3  & 0.792  & 0.400  & 49.4  & 48.5  & 47.2  & 34.6  \\ 
        ~ & DCCAE & 20.9  & 67.6  & 1.584  & 0.325  & 38.9  & 70.8  & 0.795  & 0.389  & 48.6  & 47.7  & 46.7  & 35.2  \\ 
        ~ & CPM-Net & 16.9  & 63.9  & 1.336  & 0.338  & 34.3  & 74.2  & 1.457  & 0.064  & 53.5  & 54.2  & 41.9  & 33.1  \\ 
        ~ & CRA & 31.9  & 76.2  & 1.025  & 0.660  & 48.9  & 82.7  & 0.598  & 0.691  & 56.6  & 55.7  & 54.2  & 50.9  \\ 
        ~ & MCTN & 36.0 & 75.5 & 0.918 & 0.676  & 46.8  & 78.7  & 0.642  & 0.645  & 39.6  & 41.7  & 53.3  & 46.0  \\ 
        ~ & MMIN & 38.3  & 78.6  & 0.872  & 0.701  & 50.2  & 82.0  & 0.597  & 0.693  & 52.6  & 52.7  & 55.2  & 48.1 \\ 
        ~ & GCNet & 39.0  & 81.0  & 0.832  & 0.725  & 48.1  & 81.9  & 0.622  & 0.682  & 59.2  & 59.3  & 53.2  & 51.8  \\ 
        ~ & IMDer & 39.3  & 82.8  & 0.878  & 0.729  & 49.7  & 82.2  & 0.610  & 0.694  & 59.9  & 60.3  & 55.9  & 53.9 \\ 
        ~ & LNLN & 39.9  & 79.6  & 0.837  & 0.717  & 50.1  & 82.3  & 0.591  & 0.713  & 60.5  & 61.1  & 54.2  & 52.9  \\ 
    & \textbf{CyIN} & \textbf{40.1} & \textbf{83.0} & \textbf{0.824} & \textbf{0.732} & \textbf{51.3} & \textbf{82.9} & \textbf{0.590} & \textbf{0.721} & \textbf{61.9} & \textbf{61.8} & \textbf{57.2} & \textbf{54.8} \\  
    \midrule[0.5pt]
    \multirow{11}{*}{$0.3$} ~ & CCA & 24.2  & 68.6  & 1.193  & 0.454  & 44.6  & 77.1  & 0.707  & 0.569  & 53.1  & 52.9  & 48.9  & 40.6  \\ 
        ~ & DCCA & 23.1  & 67.2  & 1.251  & 0.375  & 40.8  & 71.4  & 0.794  & 0.390  & 46.2  & 45.2  & 47.3  & 34.2  \\ 
        ~ & DCCAE & 20.4  & 66.2  & 1.545  & 0.310  & 39.1  & 69.2  & 0.801  & 0.371  & 46.3  & 45.3  & 46.9  & 34.8  \\ 
        ~ & CPM-Net & 16.9  & 64.0  & 1.336  & 0.336  & 34.1  & 74.0  & 1.180  & 0.111  & 53.0  & 53.8  & 40.4  & 34.6  \\ 
        ~ & CRA & 31.3  & 72.3  & 1.077  & 0.624  & 48.0  & 81.3  & 0.632  & 0.661  & 53.5  & 52.7  & 52.9  & 49.1  \\ 
        ~ & MCTN & 33.5  & 72.3  & 0.977  & 0.634  & 46.1  & 76.1  & 0.668  & 0.599  & 37.7  & 40.0  & 53.0  & 45.4  \\ 
        ~ & MMIN & 35.7  & 74.6  & 0.947  & 0.648  & 48.7  & 80.6  & 0.631  & 0.659  & 51.2  & 51.3  & 54.8  & 47.3 \\
        ~ & GCNet & 36.3  & 75.4  & 0.968  & 0.652  & 48.5  & 80.7  & 0.673  & 0.665  & 57.5  & 57.6  & 50.9  & 49.7  \\ 
        ~ & IMDer & 37.7  & 76.6  & 0.923  & 0.677  & 48.6  & 80.9  & 0.643  & 0.660  & 58.6  & 58.9  & 54.2  & 51.8 \\ 
        ~ & LNLN & 37.5  & 76.5  & 0.890  & 0.683  & \textbf{49.3}  & 80.8  & 0.613  & 0.683  & 59.3  & 60.2  & 52.6  & 50.9  \\ 
    & \textbf{CyIN} & \textbf{38.1} & \textbf{78.4} & \textbf{0.886} & \textbf{0.694} & 49.2 & \textbf{81.5} & \textbf{0.606} & \textbf{0.685} & \textbf{60.3} & \textbf{60.4} & \textbf{55.8} & \textbf{52.7} \\  
    \midrule[0.5pt]
    \multirow{11}{*}{$0.4$} ~ & CCA & 22.8  & 66.7  & 1.222  & 0.423  & 44.3  & 75.3  & 0.723  & 0.535  & 50.1  & 49.7  & 48.7  & 39.9  \\ 
        ~ & DCCA & 21.2  & 64.6  & 1.283  & 0.304  & 41.1  & 69.9  & 0.796  & 0.378  & 42.4  & 41.2  & 47.1  & 33.7  \\ 
        ~ & DCCAE & 19.2  & 62.9  & 1.574  & 0.273  & 39.3  & 66.6  & 0.808  & 0.347  & 43.4  & 42.2  & 46.9  & 34.1  \\ 
        ~ & CPM-Net & 16.9  & 63.8  & 1.336  & 0.335  & 34.2  & 73.9  & 1.439  & 0.061  & 52.8  & 53.6  & 41.9  & 33.2  \\ 
        ~ & CRA & 28.3  & 68.2  & 1.160  & 0.552  & 46.7  & 79.9  & 0.646  & 0.626  & 50.0  & 49.5  & 52.1  & 47.9  \\ 
        ~ & MCTN & 30.8  & 67.9  & 1.050  & 0.583  & 45.3  & 73.2  & 0.695  & 0.551  & 35.4  & 37.8  & 52.6  & 44.5  \\ 
        ~ & MMIN & 33.3  & 71.2  & 1.021  & 0.523  & 47.4  & 79.4  & 0.655  & 0.629  & 49.9  & 50.0  & 54.3  & 46.3 \\ 
        ~ & GCNet & 33.9  & 74.6  & 0.986  & 0.628  & 47.1  & 79.2  & 0.672  & 0.623  & 56.1  & 56.3  & 47.7  & 47.0  \\ 
        ~ & IMDer & 34.1  & 75.9  & 0.944  & 0.655  & 47.2  & 78.7  & 0.664  & 0.618  & 56.8  & 57.2  & 52.0  & 49.2 \\
        ~ & LNLN & 34.5  & 73.3  & 0.952  & 0.644  & 47.2  & 79.2  & 0.640  & 0.646  & 58.7  & 59.3  & 51.5  & 49.3  \\ 
    & \textbf{CyIN} & \textbf{34.7} & \textbf{76.5} & \textbf{0.935} & \textbf{0.666} & \textbf{47.5} & \textbf{79.3} & \textbf{0.635} & \textbf{0.649} & \textbf{59.3} & \textbf{59.4} & \textbf{54.4} & \textbf{50.3} \\ 
    \midrule[0.5pt]
    \multirow{11}{*}{$0.5$} ~ & CCA & 21.7  & 63.8  & 1.255  & 0.384  & 43.6  & 73.0  & 0.744  & 0.489  & 46.9  & 46.2  & 48.6  & 39.1  \\ 
        ~ & DCCA & 21.6  & 63.9  & 1.289  & 0.304  & 41.3  & 67.4  & 0.800  & 0.362  & 39.5  & 37.8  & 47.2  & 33.3  \\ 
        ~ & DCCAE & 19.5  & 60.9  & 1.558  & 0.246  & 39.3  & 63.4  & 0.818  & 0.317  & 40.1  & 38.7  & 47.4  & 34.0  \\ 
        ~ & CPM-Net & 16.9  & 63.9  & 1.335  & 0.334  & 34.2  & 74.0  & 1.626  & 0.056  & 52.7  & 53.3  & 41.1  & 33.1  \\ 
        ~ & CRA & 26.7  & 64.1  & 1.209  & 0.514  & 45.4  & 78.6  & 0.671  & 0.603  & 46.4  & 46.1  & 50.5  & 45.7  \\ 
        ~ & MCTN & 25.7  & 59.7  & 1.161  & 0.503  & 44.6  & 70.7  & 0.717  & 0.510  & 33.3  & 35.7  & 52.0  & 43.3  \\ 
        ~ & MMIN & 30.7  & 68.0  & 1.081  & 0.550  & 46.1  & 78.0  & 0.669  & 0.605  & 48.3  & 48.2  & 53.4  & 45.0 \\ 
        ~ & GCNet & 30.7  & 71.0  & 1.038  & 0.598  & 45.0  & 78.0  & 0.699  & \textbf{0.618}  & 54.3  & 54.4  & 45.6  & 45.5  \\ 
        ~ & IMDer & 31.8  & 72.2  & 0.995  & \textbf{0.605}  & 46.2  & 76.4  & 0.687  & 0.570  & 54.8  & 55.3  & 50.0  & 46.8 \\ 
        ~ & LNLN & 31.9  & 69.3  & 1.024  & 0.593  & 46.7  & 77.8  & 0.668  & 0.606  & 53.6  & 54.1  & 49.9  & 46.7  \\ 
    & \textbf{CyIN} & \textbf{32.5} & \textbf{72.6} & \textbf{0.990} & 0.599 & \textbf{47.1} & \textbf{78.7} & \textbf{0.658} & 0.611 & \textbf{55.3} & \textbf{55.5} & \textbf{53.6} & \textbf{48.2} \\ 
    \midrule[0.5pt]
    \multirow{11}{*}{$0.6$} ~ & CCA & 20.6  & 62.0  & 1.279  & 0.351  & 43.0  & 69.7  & 0.764  & 0.437  & 43.3  & 42.2  & 48.4  & 38.2  \\ 
        ~ & DCCA & 20.6  & 60.3  & 1.324  & 0.252  & 41.3  & 64.4  & 0.803  & 0.355  & 37.3  & 36.3  & 47.3  & 32.8  \\ 
        ~ & DCCAE & 19.0  & 57.4  & 1.545  & 0.231  & 39.1  & 59.2  & 0.827  & 0.293  & 37.5  & 35.2  & 47.7  & 33.4  \\ 
        ~ & CPM-Net & 17.0  & 64.0  & 1.341  & 0.299  & 34.3  & 73.9  & 1.304  & 0.093  & 52.6  & \textbf{53.2}  & 41.5  & 33.7  \\ 
        ~ & CRA & 25.1  & 61.4  & 1.264  & 0.454  & 43.9  & 77.2  & 0.696  & 0.577  & 42.7  & 42.6  & 49.6  & 44.1  \\ 
        ~ & MCTN & 24.1  & 58.8  & 1.190  & 0.464  & 44.1  & 67.4  & 0.745  & 0.452  & 30.6  & 32.9  & 51.4  & 41.8  \\ 
        ~ & MMIN & 28.3  & 62.9  & 1.158  & 0.486  & 44.8  & 76.4  & 0.688  & 0.577  & 46.3  & 46.2  & 52.1  & 43.3 \\ 
        ~ & GCNet & 29.3  & 65.9  & 1.127  & 0.513  & 43.8  & 77.2  & 0.705  & \textbf{0.584}  & 51.8  & 51.8  & 41.2  & 43.9  \\ 
        ~ & IMDer & \textbf{30.2}  & 67.8  & 1.090  & 0.564  & 44.6  & 75.3  & 0.704  & 0.544  & 52.2  & 51.7  & 47.6  & 44.8 \\ 
        ~ & LNLN & 28.9  & 64.9  & 1.150  & 0.522  & 45.6  & 76.1  & 0.689  & 0.565  & 48.1  & 47.7  & 49.3  & 45.0  \\ 
    & \textbf{CyIN} & 29.2 & \textbf{68.6} & \textbf{1.069} & \textbf{0.571} & \textbf{45.8} & \textbf{77.5} & \textbf{0.683} & 0.581 & \textbf{52.8} & 52.4 & \textbf{52.3} & \textbf{46.5} \\ 
    \midrule[0.5pt]
    \multirow{11}{*}{$0.7$} ~ & CCA & 19.9  & 59.7  & 1.298  & 0.326  & 42.7  & 68.3  & 0.775  & 0.409  & 41.2  & 39.8  & 48.5  & 37.7  \\ 
        ~ & DCCA & 19.6  & 58.6  & 1.343  & 0.214  & 41.3  & 61.9  & 0.806  & 0.360  & 36.2  & 35.0  & 47.2  & 32.5  \\ 
        ~ & DCCAE & 18.6  & 55.5  & 1.539  & 0.209  & 38.8  & 55.9  & 0.833  & 0.277  & 35.7  & 32.7  & 47.8  & 33.2  \\ 
        ~ & CPM-Net & 17.1  & 63.8  & 1.361  & 0.291  & 34.2  & 73.9  & 1.692  & 0.104  & \textbf{52.2}  & \textbf{52.5}  & 41.6  & 34.1  \\ 
        ~ & CRA & 23.6  & 57.3  & 1.311  & 0.401  & 43.0  & \textbf{76.5}  & 0.713  & 0.548  & 40.1  & 40.0  & 48.7  & 42.7  \\ 
        ~ & MCTN & 23.5  & 56.1  & 1.223  & 0.432  & 43.7  & 64.9  & 0.760  & 0.413  & 28.8  & 30.8  & 51.2  & 41.2  \\ 
        ~ & MMIN & 25.6  & 59.0  & 1.212  & 0.440  & 43.8  & 75.3  & 0.707  & 0.545  & 45.9  & 45.8  & 51.7  & 41.9 \\ 
        ~ & GCNet & 26.3  & 63.9  & 1.166  & 0.494  & 43.3  & 74.5  & 0.733  & 0.542  & 47.3  & 47.2  & 39.9  & 41.8  \\ 
        ~ & IMDer & 27.0  & 65.1  & 1.126  & 0.526  & 43.1  & 75.1  & 0.714  & 0.470  & 47.7  & 48.2  & 46.3  & 43.0 \\ 
        ~ & LNLN & 26.1  & 62.7  & 1.200  & 0.472  & 45.0  & 74.8  & 0.705  & 0.536  & 46.9  & 45.9  & 48.5  & 43.1  \\ 
    & \textbf{CyIN} &  \textbf{28.0} & \textbf{65.9} & \textbf{1.117} & \textbf{0.530} & \textbf{45.1} & 74.9 & \textbf{0.700} & \textbf{0.553} & 48.6 & 49.0 & \textbf{51.8} & \textbf{44.4} \\

    % \midrule[0.5pt]
    % \multirow{11}{*}{Avg.} &  CCA & & & & & & & & & & & &\\
    % & DCCA & & & & & & & & & & & &\\
    % & DCCAE & & & & & & & & & & & &\\
    % & CPM-Net & & & & & & & & & & & &\\
    % & CRA & & & & & & & & & & & &\\
    % & MCTN & & & & & & & & & & & &\\
    % & MMIN & & & & & & & & & & & &\\
    % & GCNet & & & & & & & & & & & &\\
    % & IMDer & & & & & & & & & & & &\\
    % & \textbf{CyIN}  & & & & & & & & & & & &\\
    \bottomrule[1.5pt]
    
\end{tabular}
}
\end{table}

The detail experiment results under various missing  modality scenarios with fixed and random missing protocols are presented in Table \ref{table_fixmiss_detail} and Table \ref{table_randommiss_detail}, respectively. Note that the reported results are the average values under 10 fixed random seeds. 

With fixed missing protocol including incomplete settings $u\in\{l\}/\{v\}/\{a\}/\{l,a\}/\{l,v\}/\{a,v\}$,  CyIN demonstrates superior or competitive performance across nearly all settings, , significantly outperforming other models in both unimodal and bimodal missing scenarios. From Table \ref{table_fixmiss_detail}, we can observe that language modality plays the most essential role in these multimodal datasets due to the rich semantics and high recognition-related information in utterance \cite{pham2019found,hazarika2020misa,lin2023dynamically}. Despite better performance in this dominant modality, CyIN pays more attention on the inferior modalities including audio and vision modalities. This result illustrates the generalization ability of the proposed framework, since CyIN has not spectacularly designed extra delicate modules to concentrate training on dominant modalities as LNLN \cite{zhang2024towards}. Besides, with one modality missing and two modalities as input, CyIN sufficiently integrate the information from bimodal interaction and reach higher performance, validating the flexibility of informative bottleneck latents in various combination of modalities.

With random missing protocol including missing rates $MR\in[0.1,0.2,0.3,0.4,0.5,0.6,0.7]$, CyIN suffers from less performance degradation as the missing rate increases. The random missing case is mostly simulating diverse missing situation in the real-world, when the presence of modalities is not fixed or known. As shown in Table \ref{table_randommiss_detail}, compared with baselines, the robustness of CyIN under various input data sparsity highlights the strength of informative bottlenecks extracted by token- and label-level IB, ensuring stable performance even when the presence of multimodal input are severely missing at $MR=0.7$. Moreover, CyIN reaches state-of-the-art performance on 4 datasets regardless of multimodal regression or classification tasks. While most baselines reach suboptimal performance on either one of the tasks due to the limitation of generalization.

The experiment results in various missing scenarios  clearly demonstrates that CyIN is highly effective in both fixed or random missing scenarios. The general  framework enables the informative space to maintain powerful performance despite the presence of input modalities, making it a robust and practical solution for real-world multimodal applications.

\subsection{Comparison of Performance Stability}
\label{appendix_stability}

\begin{table}[htbp]
\centering
\setlength\tabcolsep{7pt}
\caption{Performance stability comparison between the proposed CyIN and baselines with the average results on MOSI and IEMOCAP dataset with fixed missing protocols including modality settings $u\in\{l\}/\{a,v\}$ and random missing protocols including missing rates $MR=0.7$. }
\label{table_stablity}
\scalebox{0.8}{
    \begin{tabular}{cccccccc}
    \toprule[1.5pt]
    \multirow{2}{*}{Setting} & \multirow{2}{*}{Models} & \multicolumn{4}{c}{MOSI} & \multicolumn{2}{c}{IEMOCAP} \\
    \cmidrule(r){3-6}\cmidrule(r){7-8}
    & & Acc7$\uparrow$ & F1$\uparrow$ & MAE$\downarrow$ & Corr$\uparrow$ & Acc$\uparrow$ & wF1$\uparrow$\\

    \midrule[1.5pt]
    \multirow{7}{*}{\makecell[c]{Fixed\\{$u\in\{l\}$}}} 
        ~ & MCTN & 41.5 $\pm$ 0.71  & 83.7 $\pm$  0.37 & 0.777 $\pm$ 0.005  & 0.765 $\pm$ 0.003   & 58.4 $\pm$ 0.40  & 57.8 $\pm$ 0.41  \\ 
        ~ & MMIN & 42.3 $\pm$ 0.82  & 85.0 $\pm$ 0.45  & 0.743 $\pm$ 0.009  & 0.791  $\pm$  0.005 & 57.0 $\pm$  0.52 & 57.5 $\pm$  0.49  \\ 
        ~ & GCNet & 42.6 $\pm$ 0.53  & 85.0 $\pm$ 0.31 & 0.740 $\pm$  0.003 & 0.795 $\pm$ 0.002  & 58.7 $\pm$ 0.33  & 58.3 $\pm$  0.42  \\ 
        ~ & IMDer & 43.0 $\pm$  0.59 & 85.1 $\pm$ 0.32  & \textbf{0.717} $\pm$ 0.005  & 0.794  $\pm$ 0.002  & 60.4 $\pm$ 0.47  & 60.1  $\pm$ 0.44 \\ 
        ~ & LNLN & 43.9  $\pm$ 0.33 & 83.5 $\pm$ 0.24  & 0.761 $\pm$ 0.008  & 0.760 $\pm$ 0.006  & 62.8 $\pm$ 0.51  & 62.4 $\pm$  0.48 \\ 
        ~ & \textbf{CyIN} & \textbf{44.0} $\pm$ 0.25 & \textbf{85.2} $\pm$ 0.29 & 0.742 $\pm$ 0.004 & \textbf{0.795} $\pm$ 0.002 & \textbf{63.4} $\pm$ 0.10 & \textbf{63.0} $\pm$ 0.10 \\ 
    
    \midrule[0.5pt]
    \multirow{7}{*}{\makecell[c]{Fixed\\{$u\in\{a,v\}$}}}
        ~ & MCTN & 15.5 $\pm$ 0.71  & 42.2 $\pm$ 0.37  & 1.431 $\pm$ 0.005  & 0.018 $\pm$  0.004   & 22.7 $\pm$  0.40 & 11.5 $\pm$ 0.24   \\ 
        ~ & MMIN & 20.3 $\pm$  0.95 & 52.3 $\pm$  1.32 & 1.427 $\pm$ 0.009  & 0.081 $\pm$ 0.005  & 48.2 $\pm$ 0.51  & 47.4 $\pm$ 0.35  \\ 
        ~ & GCNet & 17.9 $\pm$ 0.83  & 57.2 $\pm$ 1.25  & 1.363 $\pm$ 0.008  & 0.385 $\pm$  0.002 & 49.0 $\pm$ 0.48 & 48.1 $\pm$  0.31   \\ 
        ~ & IMDer & 18.4 $\pm$ 0.85  & 58.1 $\pm$ 1.58  & 1.319 $\pm$  0.016 & 0.386  $\pm$ 0.026   & 50.0 $\pm$  0.30 & 49.9 $\pm$ 0.38  \\ 
         ~ & LNLN & 15.6 $\pm$ 0.75  & 48.7 $\pm$ 0.45  & 1.441 $\pm$ 0.008  & 0.076 $\pm$ 0.006  & 51.5 $\pm$ 0.56  & 50.4 $\pm$ 0.48  \\ 
    & \textbf{CyIN}  & \textbf{21.0} $\pm$ 0.35 & \textbf{59.4} $\pm$ 0.27 & \textbf{1.305} $\pm$ 0.010 & \textbf{0.391} $\pm$ 0.019 & \textbf{53.0} $\pm$ 0.40 & \textbf{52.1} $\pm$ 0.46 \\

    \midrule[0.5pt]
    \multirow{7}{*}{\makecell[c]{Random\\Missing\\$MR=0.7$}}  
        ~ & MCTN & 23.5 $\pm$ 1.15  & 56.1 $\pm$  1.23 & 1.223 $\pm$ 0.018  & 0.432  $\pm$ 0.026  & 28.8 $\pm$ 0.64  & 30.8 $\pm$ 0.69   \\ 
        ~ & MMIN & 25.6 $\pm$ 1.04  & 59.0 $\pm$ 1.60  & 1.212 $\pm$ 0.017  & 0.440 $\pm$ 0.019  & 45.9 $\pm$ 0.61  & 45.8 $\pm$  0.65 \\ 
        ~ & GCNet & 26.3 $\pm$ 2.25 & 63.9 $\pm$ 2.86 & 1.166 $\pm$ 0.038  & 0.494  $\pm$ 0.032  & 47.3 $\pm$ 0.79  & 47.2  $\pm$ 0.81  \\ 
        ~ & IMDer & 27.0 $\pm$ 1.30  & 65.1 $\pm$ 1.50  & 1.126 $\pm$  0.022 & 0.526 $\pm$  0.023   & 47.7 $\pm$  1.20 & 48.2 $\pm$ 1.12   \\ 
        ~ & LNLN & 26.1 $\pm$ 1.07  & 62.7 $\pm$ 1.63  & 1.200 $\pm$ 0.019  & 0.472 $\pm$  0.021 & 46.9 $\pm$  1.66 & 45.9 $\pm$ 1.79   \\ 
    & \textbf{CyIN} & \textbf{28.0} $\pm$ 0.90 & \textbf{65.9} $\pm$ 1.71 & \textbf{1.117} $\pm$ 0.018 & \textbf{0.530} $\pm$ 0.021 & \textbf{48.6} $\pm$ 0.71 & \textbf{49.0} $\pm$ 0.70 \\ 
    
    \bottomrule[1.5pt]
    
\end{tabular}
}
\end{table}

We evaluate the performance stability of the models on MOSI dataset under three representative circumstances, including fixed missing protocols $u\in\{l\}/\{a,v\}$ and random missing protocols $MR=0.7$ . Based on the overall average performance reported in Table \ref{table_fixmiss_detail} and Table \ref{table_randommiss_detail}, we further computed the standard deviations of the corresponding results on each missing scenarios with 10 runs, to demonstrate the performance variation degree in real-world circumstances. The final stability measure are reported in Table \ref{table_stablity}.
% when the presence of modalities can not be assured

%For the compared baselines, CPM-Nets have the lowest standard deviation across most metrics due to the extra computation in inference strategy. However, CPM-Nets remain a relative low performance on all metrics despite the missing scenarios. Other baselines either have lower performance or larger variation referring to instability of models. 

Under fixed missing modality protocols such as $u\in\{l\}/\{a,v\}$, the standard deviations of all methods show relatively small, reflecting that when the missing modalities are predetermined, models can learn to compensate in a more determine way. In contrast, random missing protocol introduce much higher variance, especially in severe missing scenario with $MR=0.7$. We summarize this into two reasons: First reason is imbalance contribution of modalities (with dominant modality like language containing more semantics compared with the inferior ones like audio or vision in specific tasks), and the second one is random missing setting is more closed to simulate the unpredictable inference circumstances in real‑world, exposing each model’s real robustness.

Across both fixed and random missing scenarios, CyIN not only achieves the superior or competitive performance on most metrics but also maintains relatively low fluctuations, highlighting the remarkable balance between performance and stability for the constructed informative space.

\subsection{Generalization Performance on Different Language Models}
To further validate the generalization performance, We train and evaluate CyIN with different sizes of Pretrained language models including BERT \cite{devlin2019bert}, RoBERTa \cite{liu2019roberta}, and DeBERTa-V3 \cite{he2023debertav3} as shown in \ref{table_different_lm}. The experiment results shows a clear benefit from scaling up the PLM backbone, as larger models like DeBERTa-V3 yield better textual representations and thus higher multimodal performance when all modalities are available. The similar trend occurs when evaluating with the most severe missing circumstance $MR=0.7$, which indicates the effective generalization ability of CyIN.

\begin{table}[htbp]
\centering
\caption{Comparison of the proposed CyIN using different sizes of language models on MOSI dataset, under both complete and randomly incomplete multimodal learning settings.}
\label{table_different_lm}
\scalebox{1.0}{
\begin{tabular}{cccccc}
\toprule
\multirow{2}{*}{Setting} & \multirow{2}{*}{\makecell[c]{Model Variants\\(PLM)}} & \multicolumn{4}{c}{MOSI} \\
\cmidrule(r){3-6}
& & Acc7$\uparrow$ & F1$\uparrow$ & MAE$\downarrow$ & Corr$\uparrow$ \\

\midrule
    \multirow{3}{*}{\makecell[c]{Complete\\{$u\in\{l,a,v\}$}}} ~ & BERT  &  48.0 & 86.3 & 0.712 & 0.801  \\
    & RoBERTa  & 49.5 & 88.2 & 0.692 & 0.823 \\ 
    & DeBERTa-V3  & \textbf{50.3} & \textbf{90.1} & \textbf{0.671} & \textbf{0.841} \\ 
    
\midrule
    \multirow{3}{*}{\makecell[c]{Random Missing\\$MR=0.7$}} ~ &  BERT  & 28.0 & 65.9 & 1.117 & 0.530 \\ 
    & RoBERTa  & 29.3 & 67.2 & 0.992 & 0.556 \\ 
    & DeBERTa-V3  & \textbf{32.2} & \textbf{69.5} & \textbf{0.965} & \textbf{0.578} \\ 

\bottomrule
\end{tabular}
}
\end{table}

\subsection{Various Scales of Other Multimodal Tasks}
We have extended experiments of CyIN on 6 non-affective datasets with 2–4 modalities and random missing rates in Table \ref{tab:additional_exp1} - \ref{tab:additional_exp4}, including the following tasks: \textbf{Multimodal Recommendation (3 modalities: visual, audio, textual)} \cite{li2025generating} (Amazon Baby, Tkitok, Allrecipes datasets), \textbf{Multimodal Face Anti-spoofing (3 modalities: RGB, Depth, IR) }\cite{wei2024robust} (CASIA-SURF dataset), \textbf{Multimodal Dense Prediction (2 modalities: RGB, Depth) } \cite{wei2024robust} (NYUv2 dataset), and \textbf{Multimodal Medical Segmentation (4 modalities: FLAIR, T1, T1c, T2)} \cite{pipoli2025imfuse} (BraTS2023 dataset). 

\begin{table}[htbp]
\centering
\setlength\tabcolsep{7pt}
\caption{Experiment comparison for \textbf{Multimodal Recommendation} task of accuracy and fairness performance (\%) on three datasets, including Amazon Baby, Tiktok and Allrecipes dataset. The modality setting is set in random missing protocol with $0.4$ missing rate as the SOTA  \cite{li2025generating} paper. }
\label{tab:additional_exp1}
\scalebox{0.77}{
% \begin{tabular}{*{14}{c}}
\begin{tabular}{c|c|cc|cc|cc|cc|cc}
\toprule
\multirow{2}{*}{Dataset} & \multirow{2}{*}{Model} &  \multicolumn{2}{c|}{Recall $\uparrow$} & \multicolumn{2}{c|}{Precision $\uparrow$} & \multicolumn{2}{c|}{NDCG $\uparrow$} & \multicolumn{2}{c|}{$F$ $\uparrow$} & \multicolumn{2}{c}{$F_{\rm{fuse}}$ $\uparrow$} \\
      &       & K=10  & K=20  & K=10  & K=20  & K=10  & K=20  & K=10  & K=20  & K=10  & K=20\\
      
\midrule
\multirow{2}{*}{Amazon Baby} & SOTA \cite{li2025generating} & 5.26 & 8.54 & 0.56 & 0.45 & 2.76 & 3.60 & 87.24 & 90.12 & 1.11 & 0.90 \\
& \textbf{CyIN(ours)} & \textbf{5.43} & \textbf{8.56} & \textbf{0.57} & 0.45 & \textbf{2.89} & \textbf{3.68} & \textbf{90.10} & \textbf{92.70} & \textbf{1.14} & 0.90 \\
\midrule

\multirow{2}{*}{Tiktok} & SOTA \cite{li2025generating} & 4.81 & 7.39 & 0.48 & 0.37 & 2.95 & 3.60 & 88.15 & 92.27 & 0.96 & 0.74 \\
& \textbf{CyIN(ours)} & \textbf{4.91} & \textbf{7.57}& \textbf{0.49}& \textbf{0.38}& \textbf{3.15}& \textbf{3.81}& \textbf{89.02}& \textbf{92.31}& \textbf{0.98}& \textbf{0.75} \\
\midrule

\multirow{2}{*}{Allrecipes} & SOTA \cite{li2025generating} & 2.49 & 3.36 & 0.24 & 0.16 & 1.33 & 1.55 & 96.82 & 89.52 & 0.49 & 0.33 \\
& \textbf{CyIN(ours)} & \textbf{2.61} & \textbf{3.43} & \textbf{0.26} & \textbf{0.17} & \textbf{1.45} & \textbf{1.66} & \textbf{99.57} & \textbf{91.94} & \textbf{0.52} & \textbf{0.34} \\

\bottomrule
\end{tabular}
}
\end{table}

\begin{table}[htbp]
\centering
\setlength\tabcolsep{7pt}
\caption{Experiment comparison for \textbf{Multimodal Face Anti-spoofing} task with Average Classification Error Rate (ACER) ($\downarrow$) metric on CASIA-SURF dataset. The modalities setting is set in fix missing protocol as the SOTA \cite{wei2024robust} paper. }
\label{tab:additional_exp2}
\scalebox{1.0}{
% \begin{tabular}{*{14}{c}}
\begin{tabular}{ccc|c|c}
\toprule
\multicolumn{3}{c|}{Modality Setting}  & \multirow{2}{*}{SOTA \cite{wei2024robust}} & \multirow{2}{*}{\textbf{CyIN(ours)}} \\
RGB & Depth & IR & & \\
\midrule
\checkmark &  &  & 7.33 & \textbf{4.48} \\
& \checkmark &  & \textbf{2.13} & 2.83 \\
 &  & \checkmark & 10.41 & \textbf{7.75} \\
\checkmark & \checkmark &  & \textbf{1.02} & 2.26 \\
\checkmark &  & \checkmark & 3.88 & \textbf{3.06} \\
 & \checkmark & \checkmark & 1.38 & \textbf{1.20} \\
\checkmark & \checkmark & \checkmark & 0.69 & \textbf{0.66} \\
\midrule
\multicolumn{3}{c|}{Average} & 3.84  & \textbf{3.18} \\
\bottomrule
\end{tabular}
}
\end{table}

\begin{table}[htbp]
\begin{minipage}[htbp]{0.45\textwidth}
\centering
\setlength\tabcolsep{7pt}
\caption{Experiment comparison for \textbf{Multimodal Dense Prediction} task with mIoU ($\uparrow$) metric on NYUv2 dataset. The modalities setting is set in fix missing protocol as the SOTA \cite{wei2024robust} paper. }
\label{tab:additional_exp3}
\scalebox{0.9}{
% \begin{tabular}{*{14}{c}}
\begin{tabular}{cc|c|c}
\toprule
\multicolumn{2}{c|}{Modality Setting}  & \multirow{2}{*}{SOTA \cite{wei2024robust}} & \multirow{2}{*}{\textbf{CyIN(ours)}} \\
RGB & Depth & & \\
\midrule
\checkmark &  & \textbf{44.06} & 43.93 \\
& \checkmark & 41.82 & \textbf{43.84} \\
\checkmark & \checkmark & 49.89 & \textbf{50.46} \\
\midrule
\multicolumn{2}{c|}{Average} & 45.26 & \textbf{46.07} \\
\bottomrule
\end{tabular}
}
\end{minipage}
\hfill
\begin{minipage}[htbp]{0.45\textwidth}
\centering
\setlength\tabcolsep{7pt}
\caption{Experiment comparison for \textbf{Multimodal Medical Segmentation} task with Dice similarity coefficient (DSC) $\%$ ($\uparrow$) metric on BraTS2023 dataset. The average results are reported conducted in fixed modality setting as the original SOTA \cite{pipoli2025imfuse} paper. }
\label{tab:additional_exp4}
\scalebox{0.9}{
% \begin{tabular}{*{14}{c}}
\begin{tabular}{c|c|c}
\toprule
Class & SOTA$^3$ & \textbf{CyIN(ours)} \\
\midrule
Enhancing Tumor& 74.6 & \textbf{75.1} \\
Tumor Core& 85.0 & \textbf{85.8} \\
Whole Tumor& 90.6 & \textbf{91.2} \\
\bottomrule
\end{tabular}
}
\end{minipage}
\end{table}

The experimental results illustrate that the proposed CyIN reaches comparable performance with the state-of-the-art (SOTA) methods, which further demonstrates the effectiveness of CyIN in generalizing to diverse numbers of modalities or multimodal fusion tasks. Moreover, the scale of datasets varies from 1,449 samples (NYU v2) to 65,671 samples (Tiktok), 87k samples (CASIA-SURF), 139,110 samples (Amazon Baby), which partially show the scaling performance of the proposed CyIN. Due to the computation resource limitation, we leave exploration on larger scale datasets in the future work.

\section{Social Impacts}
\label{appendix_social_impact}
Devoted in bridging complete and incomplete multimodal learning, the proposed CyIN shows strong potential in real-world field, such as healthcare, education, social media analysis, marketing, advertising, and human-computer interaction. By integrating data from multiple sources such as text, audio, image and video, CyIN enables multimodal system for a more detailed understanding of human emotions and intentions.

However, the use of rich and multi-source data introduces risks related to privacy violations, unauthorized surveillance, and potential misuse for manipulation. In scenarios with missing modalities, efforts to reconstruct data may introduce biases, fabricate faking information, particularly affecting underrepresented groups and raising fairness concerns. These concerns highlight the need of careful and responsible deployment for the proposed method in real-world applications.

\newpage
\section*{NeurIPS Paper Checklist}

\begin{enumerate}

\item {\bf Claims}
    \item[] Question: Do the main claims made in the abstract and introduction accurately reflect the paper's contributions and scope?
    \item[] Answer: \answerYes{} % Replace by \answerYes{}, \answerNo{}, or \answerNA{}.
    \item[] Justification: The main claims presented in the abstract and introduction clearly reflect the paper’s contributions and scope. They accurately state the design and capabilities of the proposed CyIN framework in jointly handling complete and incomplete multimodal learning.

\item {\bf Limitations}
    \item[] Question: Does the paper discuss the limitations of the work performed by the authors?
    \item[] Answer: \answerYes{} % Replace by \answerYes{}, \answerNo{}, or \answerNA{}.
    \item[] Justification: Limitations have been discussed in Section Limitations.

\item {\bf Theory assumptions and proofs}
    \item[] Question: For each theoretical result, does the paper provide the full set of assumptions and a complete (and correct) proof?
    \item[] Answer: \answerYes{} % Replace by \answerYes{}, \answerNo{}, or \answerNA{}.
    \item[] Justification: The paper have provided a complete and correct derivation proof for the mentioned equation in Appendix \ref{appendix_vib}.

    \item {\bf Experimental result reproducibility}
    \item[] Question: Does the paper fully disclose all the information needed to reproduce the main experimental results of the paper to the extent that it affects the main claims and/or conclusions of the paper (regardless of whether the code and data are provided or not)?
    \item[] Answer: \answerYes{} % Replace by \answerYes{}, \answerNo{}, or \answerNA{}.
    \item[] Justification: The paper provides detailed information on the hyperparameter settings necessary for reproducibility in Appendix \ref{appendix_hyperparam}. Moreover, as stated in the abstract, the source code is publicly released.

\item {\bf Open access to data and code}
    \item[] Question: Does the paper provide open access to the data and code, with sufficient instructions to faithfully reproduce the main experimental results, as described in supplemental material?
    \item[] Answer: \answerYes{} % Replace by \answerYes{}, \answerNo{}, or \answerNA{}.
    \item[] Justification: The datasets used in this paper are publicly accessible and available upon request. Moreover, as stated in the abstract, the source code is now publicly released, ensuring transparency and reproducibility.

\item {\bf Experimental setting/details}
    \item[] Question: Does the paper specify all the training and test details (e.g., data splits, hyperparameters, how they were chosen, type of optimizer, etc.) necessary to understand the results?
    \item[] Answer: \answerYes{} % Replace by \answerYes{}, \answerNo{}, or \answerNA{}.
    \item[] Justification: All the training and test details including dataset splits, hyper-parameters, and type of optimizer and so on are specified in Appendix \ref{appendix_hyperparam}. Besides, following previous methods, the best hyper-parameters are chosen by fifty-times of random grid search for each model.

\item {\bf Experiment statistical significance}
    \item[] Question: Does the paper report error bars suitably and correctly defined or other appropriate information about the statistical significance of the experiments?
    \item[] Answer: \answerYes{} % Replace by \answerYes{}, \answerNo{}, or \answerNA{}.
    \item[] Justification: As shown in Appendix \ref{appendix_stability}, we report the performance variation to demonstrate the superiority of the proposed method across various missing scenarios.

\item {\bf Experiments compute resources}
    \item[] Question: For each experiment, does the paper provide sufficient information on the computer resources (type of compute workers, memory, time of execution) needed to reproduce the experiments?
    \item[] Answer: \answerYes{} % Replace by \answerYes{}, \answerNo{}, or \answerNA{}.
    \item[] Justification: We report the required computer resources in Implementation Details of Section \ref{experiments}.
    
\item {\bf Code of ethics}
    \item[] Question: Does the research conducted in the paper conform, in every respect, with the NeurIPS Code of Ethics \url{https://neurips.cc/public/EthicsGuidelines}?
    \item[] Answer: \answerYes{} % Replace by \answerYes{}, \answerNo{}, or \answerNA{}.
    \item[] Justification:  This research conforms all the requirements of moral and ethical norms in the NeurIPS Code of Ethics.

\item {\bf Broader impacts}
    \item[] Question: Does the paper discuss both potential positive societal impacts and negative societal impacts of the work performed?
    \item[] Answer: \answerYes{} % Replace by \answerYes{}, \answerNo{}, or \answerNA{}.
    \item[] Justification: We have discussed both positive and negative social impacts in Appendix \ref{appendix_social_impact}.
    
\item {\bf Safeguards}
    \item[] Question: Does the paper describe safeguards that have been put in place for responsible release of data or models that have a high risk for misuse (e.g., pretrained language models, image generators, or scraped datasets)?
    \item[] Answer: \answerNA{} % Replace by \answerYes{}, \answerNo{}, or \answerNA{}.
    \item[] Justification: The paper utilizes public used datasets and pre-trained models, which poses no such risks.

\item {\bf Licenses for existing assets}
    \item[] Question: Are the creators or original owners of assets (e.g., code, data, models), used in the paper, properly credited and are the license and terms of use explicitly mentioned and properly respected?
    \item[] Answer: \answerYes{} % Replace by \answerYes{}, \answerNo{}, or \answerNA{}.
    \item[] Justification:  The used data have been correctly cited in the paper and the license and terms of use are properly respected.

\item {\bf New assets}
    \item[] Question: Are new assets introduced in the paper well documented and is the documentation provided alongside the assets?
    \item[] Answer: \answerYes{} % Replace by \answerYes{}, \answerNo{}, or \answerNA{}.
    \item[] Justification: As shown in Abstract, the source code is publicly released.

\item {\bf Crowdsourcing and research with human subjects}
    \item[] Question: For crowdsourcing experiments and research with human subjects, does the paper include the full text of instructions given to participants and screenshots, if applicable, as well as details about compensation (if any)? 
    \item[] Answer: \answerNA{} % Replace by \answerYes{}, \answerNo{}, or \answerNA{}.
    \item[] Justification: The paper does not involve crowdsourcing nor research with human subjects.

\item {\bf Institutional review board (IRB) approvals or equivalent for research with human subjects}
    \item[] Question: Does the paper describe potential risks incurred by study participants, whether such risks were disclosed to the subjects, and whether Institutional Review Board (IRB) approvals (or an equivalent approval/review based on the requirements of your country or institution) were obtained?
    \item[] Answer: \answerNA{} % Replace by \answerYes{}, \answerNo{}, or \answerNA{}.
    \item[] Justification: The paper does not involve crowdsourcing nor research with human subjects.

\item {\bf Declaration of LLM usage}
    \item[] Question: Does the paper describe the usage of LLMs if it is an important, original, or non-standard component of the core methods in this research? Note that if the LLM is used only for writing, editing, or formatting purposes and does not impact the core methodology, scientific rigorousness, or originality of the research, declaration is not required.
    %this research? 
    \item[] Answer: \answerNA{} % Replace by \answerYes{}, \answerNo{}, or \answerNA{}.
    \item[] Justification: The core method development in this research does not involve LLMs as any important, original, or non-standard components.

\end{enumerate}

\end{document}